\documentclass[a4paper,fleqn]{cas-dc}

\usepackage[numbers]{natbib}
\usepackage{comment}
\usepackage{balance}
\usepackage{graphicx}
\usepackage{todonotes}
\usepackage{color}
\usepackage{soul}
\usepackage[super]{nth}
\usepackage{adjustbox}

\usepackage[ruled, lined, linesnumbered, commentsnumbered, longend]{algorithm2e}
\usepackage{xcolor}

\def\tsc#1{\csdef{#1}{\textsc{\lowercase{#1}}\xspace}}
\tsc{WGM}
\tsc{QE}
\tsc{EP}
\tsc{PMS}
\tsc{BEC}
\tsc{DE}
 \usepackage{subfigure}
 \usepackage{multirow}
\usepackage{array}
\usepackage{tabularx}
\newcolumntype{P}[1]{>{\centering\arraybackslash}p{#1}}
\usepackage{hhline}
\usepackage{colortbl}
\usepackage{soul}
\usepackage{todonotes}

\usepackage{soul}
\usepackage{todonotes}

\newcommand\scalemath[2]{\scalebox{#1}{\mbox{\ensuremath{\displaystyle #2}}}}

\usepackage{soul}
\usepackage{todonotes}

\begin{document}
\let\WriteBookmarks\relax
\def\floatpagepagefraction{1}
\def\textpagefraction{.001}
\shorttitle{AI-Powered Cow Detection in Complex Farm Environments}
\shortauthors{Voncarlos, Rili, Gisiger, Gambs, Vasseur, Cellier and Diallo.}

%\title [mode = title]{This is a specimen $a_b$ title}
\title [mode = title]{AI-Powered Cow Detection in Complex Farm Environments}

%\tnotemark[1,2]

%\tnotetext[1]{This document is the results of the research
  % project funded by the National Science Foundation.}

%\tnotetext[2]{The second title footnote which is a longer text matter
  % to fill through the whole text width and overflow into
   %another line in the footnotes area of the first page.}

\author[1,3]{Voncarlos M. Araújo}%[type=editor,
                        %auid=000,bioid=1,  %orcid=0000-0001-7511-2910]
%\cormark[1]
%\ead{cvr_1@tug.org.in}
%\ead[url]{www.cvr.cc, cvr@sayahna.org}
%\credit{Conceptualization of this study, Methodology, Software}

\address[1]{Département d’informatique, Université du Québec à Montréal, Montréal, Québec, Canada}

\address[2]{McGill University, Montréal, Québec, Canada}
\address[3]{Innovation Chair in Animal Welfare and Artificial Intelligence (WELL-E)}

\author[1,3]{ Ines {Rili}}
\author[1,3]{ Thomas {Gisiger}}
\author[1,3]{ Sébastien {Gambs}}
\author[2,3]{ Elsa {Vasseur}}
\author[2,3] {
Marjorie {Cellier}}
\author[1,3]{ Abdoulaye Baniré {Diallo}}

%\credit{Data curation, Writing - Original draft preparation}
%\address[3]{STM Document Engineering Pvt Ltd., Mepukada,Malayinkil, Trivandrum 695571, India}
%\cortext[cor2]{Principal corresponding author}
%\fntext[fn1]{This is the first author footnote. but is common to third
 % author as well.}
%\fntext[fn2]{Another author footnote, this is a very long footnote and
 % it should be a really long footnote. But this footnote is not yet
 % sufficiently long enough to make two lines of footnote text.}

%\nonumnote{This note has no numbers. In this work we demonstrate $a_b$
  %the formation Y\_1 of a new type of polariton on the interface
  %between a cuprous oxide slab and a polystyrene micro-sphere placed
  %on the slab.
  %}

\begin{abstract}
Animal welfare has become a critical issue in contemporary society, emphasizing our ethical responsibilities toward animals, particularly within livestock farming. 
In addition, the advent of Artificial Intelligence (AI) technologies, specifically computer vision, offers a innovative approach to monitoring and enhancing animal welfare.
Cows, as essential contributors to sustainable agriculture and climate management, being a central part of it. 
However, existing cow detection algorithms face significant challenges in real-world farming environments, such as complex lighting, occlusions, pose variations and background interference, which hinder accurate and reliable detection. 
Additionally, the model generalization power is highly desirable as it enables the model to adapt and perform well across different contexts and conditions, beyond its training environment or dataset. 
This study addresses these challenges in diverse cow dataset composed of six different environments, including indoor and outdoor scenarios. 
%Seb: below put the meaning of the CBAM acronym
%Voncarlos: ok
More precisely, we propose a novel detection model that combines YOLOv8 with the CBAM (Convolutional Block Attention Module) and assess its performance against baseline models, including Mask R-CNN, YOLOv5 and YOLOv8. 
Our findings indicate that while baseline models show promise, their performance degrades in complex real-world conditions, which our approach improves using the CBAM attention module.
Overall, YOLOv8-CBAM outperformed YOLOv8 by 2.3\% in mAP across all camera types, achieving a precision of 95.2\% and an mAP@0.5:0.95 of 82.6\%, demonstrating superior generalization and enhanced detection accuracy in complex backgrounds.
Thus, the primary contributions of this research are: (1) providing an in-depth analysis of current limitations in cow detection under challenging indoor and outdoor environments, (2) proposing a robust general model that effectively detects cows in complex real-world conditions and (3) evaluating and benchmarking state-of-the-art detection algorithms. 
Potential application scenarios of the model include automated health monitoring, behavioral analysis and tracking within smart farm management systems, enabling precise detection of individual cows, even in challenging environments.
By addressing these critical challenges, this study paves the way for future innovations in AI-driven livestock monitoring, aiming to improve the welfare and management of farm animals while advancing smart agriculture.
\end{abstract}

%\includegraphics{figs/grabs.pdf}

\begin{comment}

\begin{highlights}

\item Developed YOLOv8-CBAM model for improved cow detection accuracy.

\item Cow dataset includes six environments, addressing occlusions and lighting issues.

\item Model generalizes well in both indoor and outdoor scenarios.

\item YOLOv8-CBAM outperformed YOLOv8 by 2.3\% in mAP@0.5:0.95.

\item Enhanced precision of 95.2\% achieved across complex farm conditions.

\end{highlights}
\end{comment}

\begin{keywords}
Animal welfare \sep
Cow detection \sep
Livestock Monitoring \sep
Deep Learning \sep
%Computers and Electronics in Agriculture \sep
%Sensors \sep

\end{keywords}

\maketitle

\section{Introduction}

Animals are vital components of our ecosystem, living alongside us in a complex network of life~\cite{Mellor2017}. 
As creatures capable of feeling and awareness, they merit our care and vigilant consideration for their well-being. 
The significance of animal welfare has become increasingly acknowledged, reflecting our growing awareness of the ethical and moral responsibilities linked to our interactions with animals~\cite{Mellor2017, Clark2016ASR}. 
Thus, whether they are farm animals, domestic pets, wild animals or subjects in scientific research, guaranteeing their welfare is now a essential concern in contemporary society \cite{Wang2019MachineLF,brito2020large}.
At the same time, the domain of Artificial Intelligence (AI) has seen remarkable advances, covering machine learning, neural networks, and deep learning, exhibiting significant performance in multiple domains \cite{sharifani2023machine,talaei2023deep}. 
One of the most promising applications of AI lies in its integration into the field of livestock monitoring \cite{borger2021,Bao2022}, where there is a unique opportunity to modernize the surveillance, administration and advancement of animal welfare in various contexts \cite{gehlot2022dairy}.

Within the realm of livestock farming, the welfare of cows has attracted considerable attention for several critical reasons \cite{phillips2008cattle}. 
%Seb: for each of this point it would be great to include a reference for the claim being made
%Voncarlos: ok
First, cows play a critical role in sustainable climate management, as well-managed pastures with cows can sequester carbon, capture carbon dioxide from the atmosphere and store it in the soil \cite{schwartz2013cows}. Ensuring their health and well-being is therefore essential for combating climate change globally. 
Second, ethical considerations regarding animal treatment have increasingly influenced consumer preferences and regulatory standards, leading to increased inspection and investment in improving livestock conditions \cite{matheny2007farm}. 
Third, scientific research continues to underscore the correlation between animal welfare and productivity, demonstrating that healthier, happier cows can lead to higher quality products and more efficient farming practices \cite{buller2018towards}. 
Finally, advances in technology, such as AI-powered monitoring systems, have provided new avenues to effectively monitor and enhance cow welfare, driving greater attention and innovation in this field. 

Most of the current research on cow welfare has been conducted into two main categories: wearable sensors and non-contact technologies~\cite{Li2023}. 
Wearable sensors collect data on body temperature, sound, pulse and activity levels to infer behavior and physiological status~\cite{Barker2018, Benaissa2019}. 
However, this method incurs high material costs and has limited battery life, impacting data reliability and contributing to landfill waste. 
In contrast, purely computer vision-based cow recognition has advanced with the rise of smart agriculture and labor shortages~\cite{Mar2023, Xiao2022}, this method being both non-invasive and efficient. 
For example, \citet{Hao2023} have proposed a cattle body detection approach using YOLOv5-EMA, incorporating the Efficient Multi-scale Attention (EMA) module to improve small-object detection, achieving an average precision of~95.1\%.
Nonetheless, the model's performance in real-world scenarios still needs improvement. 
In particular in practical farming settings, current cow detection algorithms face significant limitations as they often struggle with increased noise and perform poorly under diverse and complex lighting conditions, including strong light, low light and varying levels of density. 
This underscores the ongoing necessity for enhancing algorithm accuracy, which limits the robustness of detection in complex settings.

\begin{figure}
  \centering
  \includegraphics[width=\linewidth]{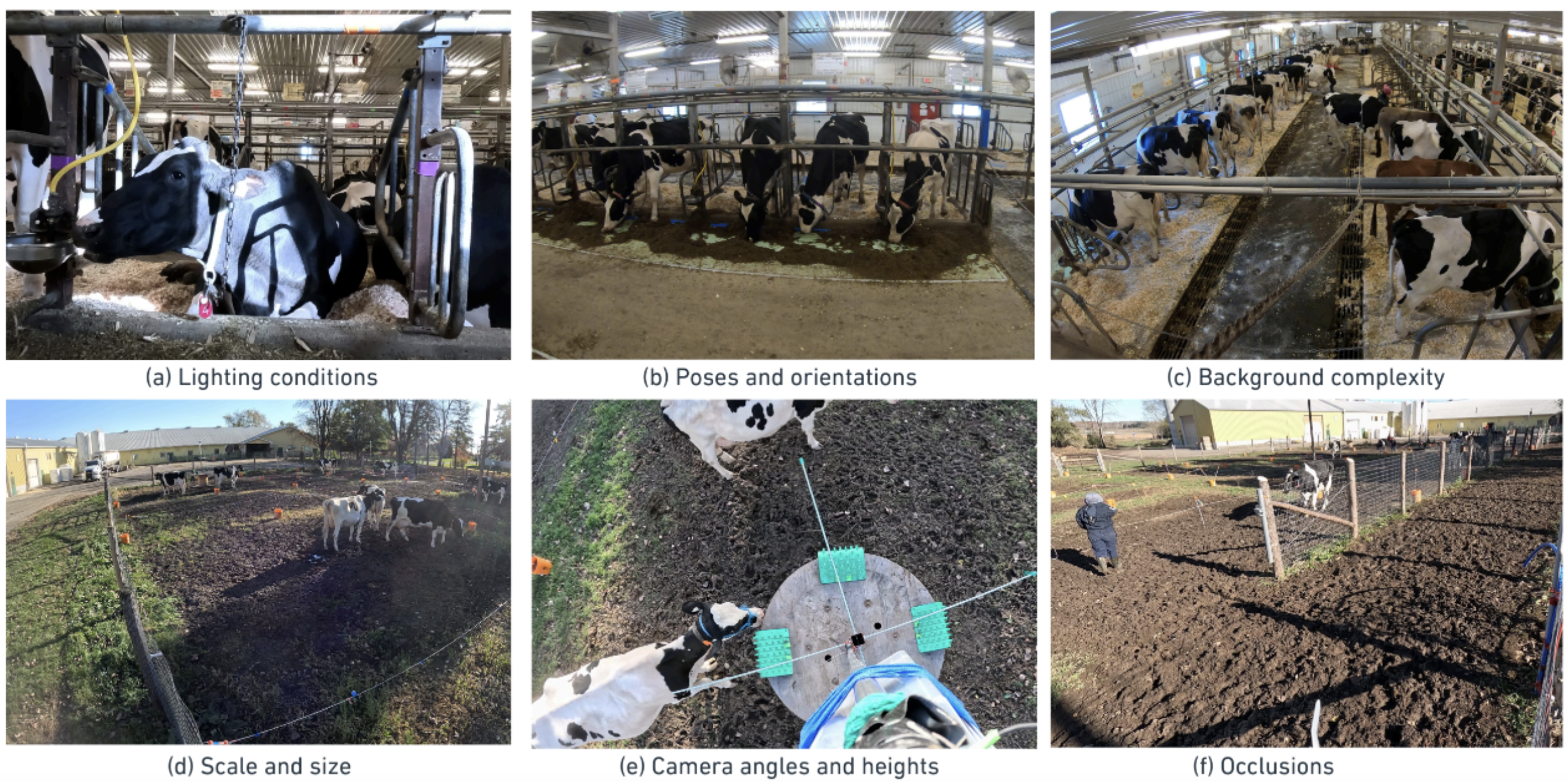}
  \caption{Cow dataset challenges: Indoors: (a) lighting conditions, (b) poses and orientations and (c) Background complexity. Outdoors: (d) scale and size, (e) camera angles and heights and (f) occlusions.}
  \label{EnsembleAttack4}
\end{figure}

To address these challenges, our study has first consisted in curating a cow dataset that covers six diverse environments, including both indoor and outdoor scenarios.
Figure~\ref{EnsembleAttack4} showcases the indoor (top row) and outdoor (bottom row) cameras, illustrating various challenges in the proposed dataset, such as (a) lighting conditions, (b) pose and orientation, (c) background complexity, (d) scale and size, (e) camera angles and heights and (f) occlusions. 
These scenarios are frequently encountered in real world due to factors such as: (1) animals frequently change posture when moving between outdoors and indoors; (2) high similarity in coat color among animals of the same species; (3) variable lighting conditions throughout the day, with low light in the morning and strong shadows or over-illumination at noon, which can lead to misinterpretation of patches or shadows as animal features; (4) complex backgrounds, including social interactions, cow gatherings and the ground environment, which can make it difficult to distinguish segmented cows from their surroundings, with some instances nearly blending into the soil.

To solve these issues, we have employed and compared two-stage models as the Mask R-CNN and one-stage YOLOv5 and YOLOv8 detection algorithms. 
In addition, we have assessed their performance in complex environments and suggest enhancements, particularly with YOLOv8 integrated with the Convolutional Block Attention Module (CBAM). 
The CBAM attention mechanism module enhances detection accuracy by focusing on key areas and reducing background noise. 
More precisely, it computes attention weights for both spatial and channel dimensions, refining input features through these weights. 
This adaptation aims to dynamically learn critical features, enhancing the average accuracy of cow detection while minimizing instances of missed detections, false detections, and low confidence.

Given the aforementioned issues, the main contributions this study can be summarized as follows:
%key contributions:  
\begin{itemize}
     \item We have compiled a cow dataset of approximately 1,115 images, with around 871 captured indoors on the farm using three different cameras and about 244 images taken outside with three additional cameras. 
     The dataset addresses several challenges in cow detection, including varying lighting conditions, diverse backgrounds, pose variations, differences in cow sizes and distances and environmental obstacles such as occlusions, partial visibility and clutter as demonstrated in Figure~\ref{EnsembleAttack4}.
     \item  We have employed both one-stage and two-stage detection models to determine the best approach for cow dataset. 
     We have also conducted a variety of experiments to compare detection performance between the baseline detection model and the proposed method enhanced with the CBAM attention mechanism. 
     This involved assessing various improvements, visualizing the effects of these modifications and evaluating the overall performance of the final model. 
    \item The integration of the CBAM substantially improves detection performance under challenging conditions, including small-scale targets, shadows and occlusions. 
    An in-depth analysis of each camera individually shows that CBAM has achieved notable precision–recall (P–R) curve improvements, with increases of 8.1\%, 13.4\%, and 2.8\% respectively for the challenging cameras IC, OE and IP, compared to the baseline model under these conditions.
    \item We have validated the proposed model’s generalization power with a unique design that detects cows in both indoor and outdoor environments. 
    It effectively identifies animals against complex backgrounds and remains resilient to environmental factors and anomalies such as occlusions and partial visibility, ensuring adaptability for various real-world applications. 
    \item With respect to the cow dataset, our proposed YOLOv8-CBAM model achieves a precision of 95.2\% and an mAP@0.5:0.95 of 82.6\%, which are improvements of 2.1\% and 1.8\%, respectively, over the original YOLOv8 model.
 \end{itemize}

The outline of the paper is as follows. 
%Seb: to complete with a reference to the corresponding sections
%Voncarlos: ok
First, in Section~\ref{sect_related_work}, we introduce the related work and highlight the existing challenges in dairy cow detection. 
%Seb: the sentence below does not seem to match the current outline
%Voncarlos: ok
%Next, in Section ??? we present the one-stage and two-stage detection models used for cow detection, discussing the advantages and disadvantages of the baseline models while providing background and context for the proposed improvements. 
Afterwards in Section~\ref{sect_materials_methods}, we cover the materials and methods of the cow data collection process, followed by data augmentation techniques and data labeling. 
Additionally, the proposed YOLOv8-CBAM detection approach is compared with traditional detection algorithms. 
The integration of the CBAM attention module is introduced to improve feature extraction in both complex indoor and outdoor environments. 
In Section~\ref{sect_experiments}, we describe the experimental validation conducted for evaluating the performance of the improved YOLOv8-CBAM algorithm against the baseline model in terms of accuracy and robustness. 
Then, we present and analyze the results in Section~\ref{comparisonMethods}, demonstrating the effectiveness of the proposed approach. 
Finally, in Section \ref{sect_discussion}, we discuss the proposed approach, including a review of its advantages, recognition of its limitations, and suggestions for future research and applications before concluding in Section \ref{sect_conc}.

\section{Related Work}
\label{sect_related_work}

Traditional cow detection methods typically combine techniques such as color segmentation~\cite{bello2021enhanced}, texture analysis \cite{hu2020cow} and shape feature analysis \cite{yao2019cow}. 
However, these approaches are often sensitive to external conditions, leading to reduced detection accuracy in complex environments. 
In contrast, deep learning-based methods automatically learn feature representations from data, eliminating the need for manual feature engineering and providing greater flexibility to handle challenging image conditions. 
When trained on large datasets, these models adapt effectively to various environmental factors, including lighting variations and particulate noise, thus demonstrating robustness in complex situations. 

Deep learning detection algorithms can be categorized into one-stage and two-stage methods. 
One-stage detectors, such as the YOLO series \cite{redmon2018yolov3, bochkovskiy2020yolov4, yolov55} and SSD \cite{liu2016ssd}, offer speed and efficiency by outputting class and location information in a single step. In object detection tasks, the class corresponds to the type of object being detected, such as 'person,' 'vehicle,' or 'animal,' while the location refers to the bounding box coordinates identifying the object's position in the image.
%Seb: with respect to the different tasks, I suggest to provide more information about the prediction task (i.e., what do the class correspond to). 
%Voncarlos: I added the following sentence: In object detection tasks, the class corresponds to the type of object being detected, such as 'person,' 'vehicle,' or 'animal,' while the location refers to the bounding box coordinates identifying the object's position in the image.
In contrast, two-stage detectors generate candidate regions before classifying and refining these regions, which typically results in higher accuracy. 
Notable examples include Fast R-CNN \cite{girshick2015fast}, Mask R-CNN \cite{he2018maskrcnn} and Cascade R-CNN \cite{cai2018cascade}.

In the context of two-stage animal target detection algorithms, \citet{andrew2021visual} have applied the Faster R-CNN model \cite{RenHG015} to annotate and train a dataset of aerial images featuring cows, achieving a remarkable mean Average Precision (mAP) of 99.6\%. 
Despite this high mAP, detecting this breed is relatively easier due to the significant contrast between their body coloration and the background. 
However, it is essential to note that the Faster R-CNN model generally requires longer training times compared to single-stage models. 
\citet{xu2020automated} utilized a quadcopter to capture images of cows in various settings, including open pastures and fenced feeding areas. 
By applying the Mask R-CNN model \cite{he2018maskrcnn} for labeling and training, they achieved an mAP of 96\% for cow body detection in open pastures and 94\% in feeding grounds. 
Similarly, \citet{xiao2022cow} employed an enhanced Mask R-CNN model on a barn dataset of cows, achieving an impressive 97.39\% mAP. 
However, like Faster R-CNN, these methods belong to the two-stage category, resulting in slower training speeds.

For the one-stage cow target detection algorithms, \citet{tassinari2021computer} used the YOLOv3 model \cite{redmon2018yolov3} with a stationary camera setup to capture and annotate indoor images of cows, reporting an average detection accuracy of 64\%. 
Likewise, \citet{lodkaew2023cowxnet} have installed fixed cameras on cow farms to collect and label cow datasets, training the data with the YOLOv4 model \cite{bochkovskiy2020yolov4} and achieving over 90\% mAP in detection accuracy. Collectively, these studies highlight the potential of deep learning approaches for dairy cow detection. 
However, a significant research gap remains with respect to achieving dairy cow detection in complex scenarios.

%\avtd{We need to decide whether to place the literature review: in this section or in the introduction section, and identify which topics to explore. I was considering focusing on one-stage and two-stage detection models in animal welfare, particularly since there are limited studies available on cows in the existing literature.}

%\subsection{Cow Detection using IA}
%\subsubsection{Two-stage Detection Models}
%\subsubsection{One-stage Detection Models}

\section{Materials and Methods}
\label{sect_materials_methods}

\begin{figure*}
% \begin{adjustwidth}{-\extralength}{0cm}
\centering
\includegraphics[width=18cm]{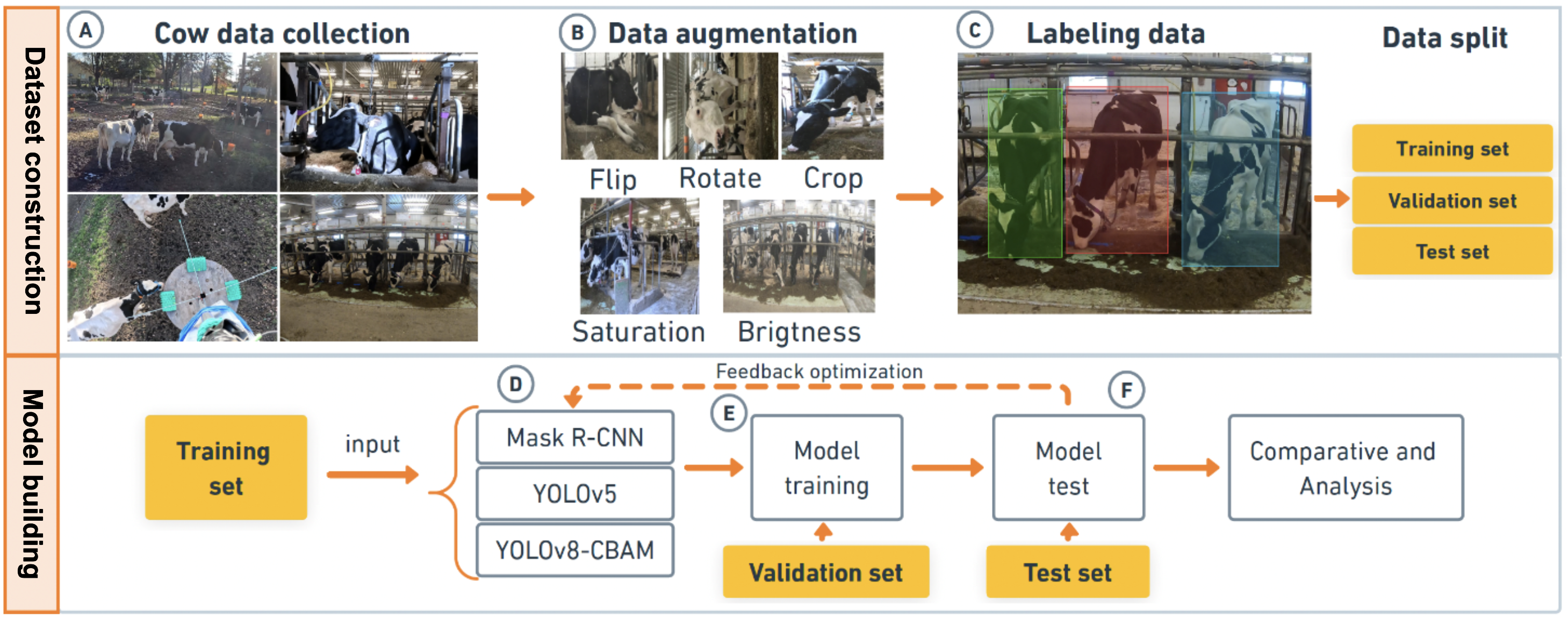}
% \end{adjustwidth}
\caption{Pipeline Overview. A) Cow data collection; B) Data augmentation; C) Labeling data; D) Detection models; E) Model training and F) Model test and Comparative/Analysis}
\label{proposedMethod}
\end{figure*}

The proposed pipeline for detecting cows in complex environments is illustrated in Figure \ref{proposedMethod}. 
First, for dataset construction, images of dairy cows were collected in various environments. 
Then, data augmentation techniques were applied to increase the diversity and quantity of images, followed by the creation of annotation files necessary for model training. 
The dataset was then split into training (70\%), validation (15\%), and testing (15\%) subsets for evaluation. 
%Seb: above I suggest to put the percentage used for the splitting
%Voncarlos:ok
Subsequently, the two-stage Mask R-CNN, one-stage YOLOv5 and the adaptive YOLOv8-CBAM models were trained and tested. 
Finally, the detection results are compared with those obtained prior to the adaptations. 
The details of each component of the pipeline are described in the following sections.

\subsection{Data Collection}

%\avtd{I think we should expand this section by adding more details about the farm, including how and when the pictures were captured, camera positions, and other relevant information. Adding an image of the camera setup would also be great. I believe someone from McGill has more expertise to complete this section.}

Our study utilized a total of 1,115 cow images captured in two field environments: indoors and outdoors, as illustrated in Figure \ref{EnsembleAttack4}. 
These images were taken at the Macdonald Campus Farm located in Ste-Anne de Bellevue,  Québec, Canada. 
All images were shot using the cameras AXIS M3057-PLR, boasting a resolution of 1920 $\times$ 1080 pixels and saved in the JPG image format. 
During the photography process, the indoors cameras were positioned approximately 0.96cm–4m from the cows and outdoors in a pole with 4 meters in the corner of the farm.

\subsection{Image Processing and Data Augmentation}

The collected images were adjusted to a resolution of 640 $\times$ 640 to conform to the input requirements of the detection model. 
To improve the model's training effectiveness, the cow image dataset was augmented using various image enhancement techniques. 
More precisely, different transformation methods were applied: mirror flip, rotation, random cropping, saturation and brightness, as depicted in Figure~\ref{proposedMethod} (b). 
The dataset numbers after data augmentation are shown in the ``Train Set Augmented'' column of Table \ref{tabledataset}. 
Notably, we did not balance the dataset post-augmentation, as one of our objectives is to evaluate the model's generalization power, even with limited images for certain camera types.

%Seb: we need to be sure that the acronym of the type of cameras are detailed somewhere
\small
\begin{table}[width=.9\linewidth,cols=3,pos=h]
\caption{The total number of training images per cameras. 
}\label{tabledataset}
\begin{tabular}{@{}p{0.2cm}ccccc|c@{}}
\toprule
  \parbox[t]{2mm}{\multirow{2}{*}{\rotatebox[origin=c]{90}{}}}& Type  &  Number & Train   & Valid& Test & Train Set\\
 & Camera & Images & Set   & Set & Set& Augmented\\ \midrule
\parbox[t]{2mm}{\multirow{3}{*}{\rotatebox[origin=c]{90}{Indoor}}}  & IP  & 240   & 181    & 36  & 23    & 539     \\
 & IW                     & 603   & 429   & 111  & 63 & 1,397             \\
 & IC                  & 28   & 19   & 6  & 3   & 66           \\
 \midrule
 \parbox[t]{2mm}{\multirow{3}{*}{\rotatebox[origin=c]{90}{Outdoor}}} & OE                    & 39      & 27     & 7  & 5       & 81    \\
 & OC                         & 58    & 40    & 12  & 6     & 169        \\
 & OP                      & 147    & 107     & 20 & 20 & 321             \\
\bottomrule
Total & - & 1,115 & 803 & 192 & 120 & 2,340 \\ \midrule
\end{tabular}
\end{table}

\subsection{Data Labeling}

For the annotation technique shown in Figure \ref{proposedMethod} (c), we utilize the same approach as \citet{automaticdetection} to improve annotation efficiency, dividing the process into two phases: manual and semi-manual box annotation. 
The goal is to reduce the time needed for manual annotation by first developing an automated baseline detector model. 
This model acts as the initial step, followed by updating subsequent annotated images with the proposed model. 
To achieve this, we start by randomly selecting images from the dataset, manually annotating a subset of cow images, and using these annotations to train a baseline cow detector. 
This detector then generates predictions, which we manually correct to ensure accuracy.

\begin{itemize}
    \item \emph{Manual cow box annotation.} We select randomly 600 cow images from the dataset and annotate them using the Roboflow Annotator toolkit.
    %Seb: put a reference for the annotator toolkit
    These images are split into 500 for training and 100 for validation. 
    We then train a baseline cow detector on this subset to prepare for the next phase.
    \item \emph{Semi-manual cow box annotation}. This phase integrates machine-learning predictions with human refinements. 
    The baseline detector generates candidate boxes on cow images, which are then reviewed by a human expert. The expert corrects false positives (\emph{e.g.}, distractors like inclusions or shadows) and adds any missed cows.
\end{itemize}

Following the fast and efficient annotation process, the 1115 cow images were randomly divided into three sets: training, validation and testing, maintaining a ratio of 7:2:1, as detailed in Table \ref{tabledataset}.

\subsection{Detection Models}

In Figure \ref{proposedMethod} (d), we propose employing both the one-stage YOLO version and the two-stage Mask R-CNN detectors, alongside the integration of attention mechanisms via YOLOv8-CBAM. 
These components are described in detail in this section.

\subsubsection{Mask R-CNN}

\begin{figure}
  \centering
  \includegraphics[width=\linewidth]{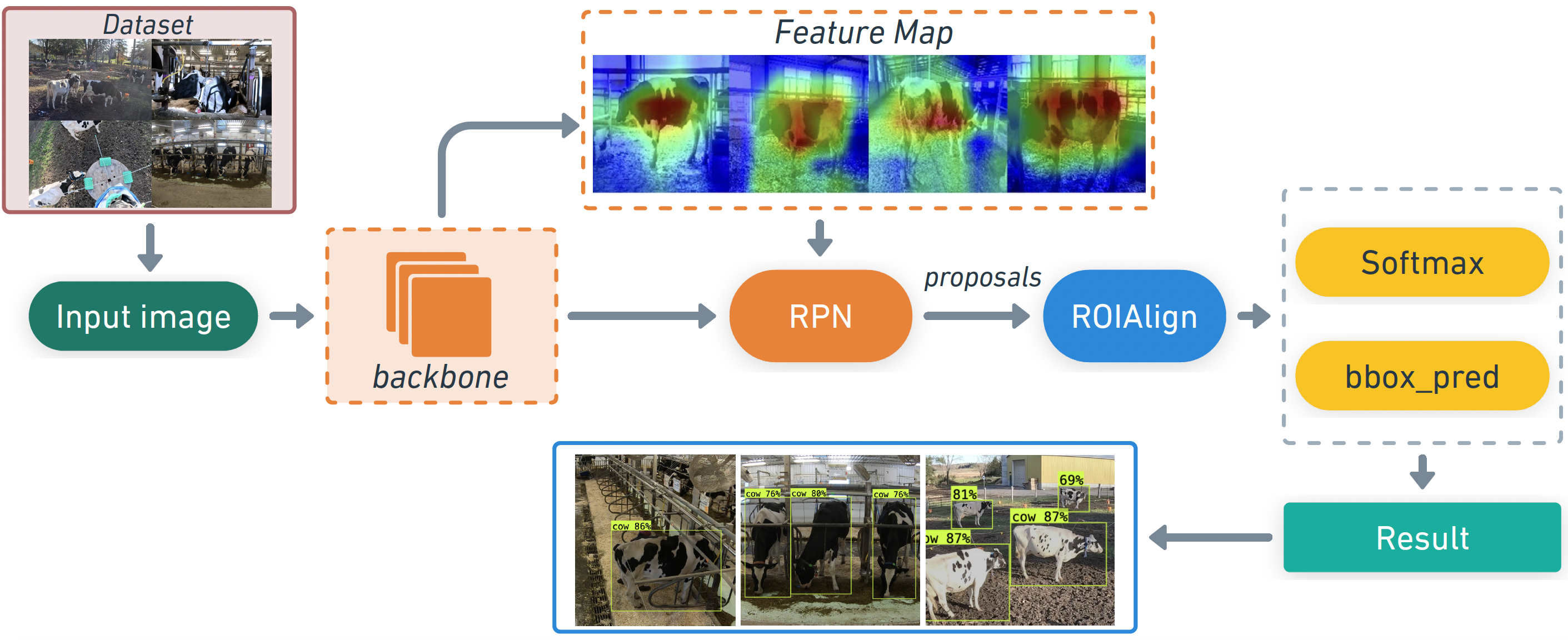}
  \caption{Framework of Mask R-CNN based cow detection.}
  \label{EnsembleAttack3}
\end{figure}

The Mask R-CNN is a two-stage instance segmentation algorithm, and the first
one was proposed by \citet{he2018maskrcnn} in 2017.
Figure \ref{EnsembleAttack3} illustrates the framework for Mask R-CNN-based cow detection, which consists of two main components: (1) Convolutional Backbone: responsible for extracting features from the entire image; (2) Head: handles bounding box recognition (classification and regression) and mask prediction. 
The Region Proposal Network (RPN) generates region proposals, after which RoIAlign extracts features from each proposal and performs two parallel tasks: one for cow detection, classification and bounding box regression through fully connected layers, and the other for producing high-accuracy segmentation masks via RoIAlign.

The features extracted by RoIAlign are input into the Fully Connected (FC) layer for classification and bounding box regression and also into the convolutional layer for segmentation. 
Classification is performed by passing the output from the FC layer, which incorporates all features, through a softmax layer. 
For bounding box regression, only the features from the original region proposal are used. 
Meanwhile, the Mask R-CNN head predicts segmentation masks for the cow body regions. 
During network training, the loss function measures the discrepancy between the predicted values and the ground truth. 
Thus, this function is crucial for training the cow segmentation model. 
In our Mask R-CNN model, a joint loss function is employed to train the bounding box regression, classification and mask prediction branches, which is detailed in \citet{he2018maskrcnn}.

In summary, for each cow image, CNN features are first extracted using ResNet-101 as backbone. 
The RPN then employs a sliding window approach \citet{RenHG015} on the feature maps to compute bounding box proposals. 
RoIAlign is used to map regions of interest with varying sizes in the feature maps to a fixed spatial resolution using bilinear interpolation. 
Finally, the Mask R-CNN head predicts object classes, refines bounding box localization, and generates segmentation masks simultaneously.

\subsubsection{YOLOv5}

You Only Look Once (YOLO), first introduced by \citet{yolov55}, has revolutionized object detection with its end-to-end network that can simultaneously detect object bounding boxes and classify their labels. 
Unlike the R-CNN series algorithms, the YOLO series integrates object localization and recognition into a single step, allowing it to directly generate bounding box coordinates and class predictions from a single neural network regression. 
This innovative approach simplifies the network structure, significantly boosts detection efficiency, and maintains reasonable accuracy, making it well-suited for real-time applications.

\begin{figure}
  \centering
  \includegraphics[width=\linewidth]{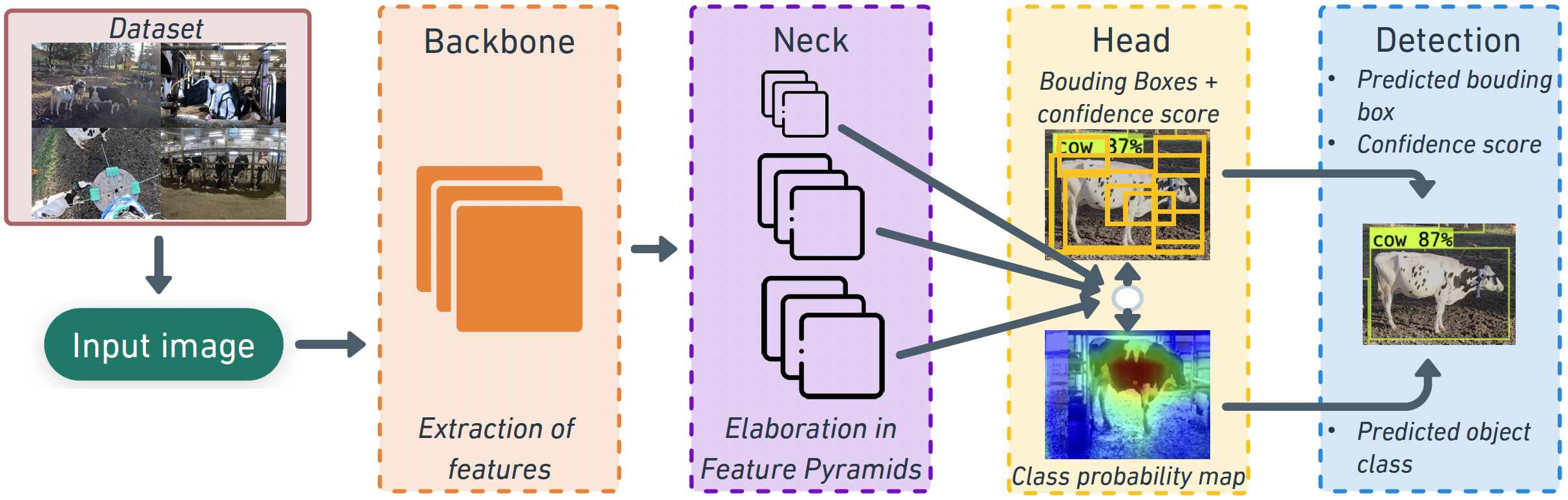}
  \caption{YOLOv5 detection model.}
  \label{yoloarch}
\end{figure}

As seen in Figure \ref{yoloarch}, YOLOv5 model comprises three main components: backbone, neck and head \cite{yolov5}. 
The backbone of the network is primarily responsible for extracting image features at various levels and includes modules such as CBS, C3 and spatial pyramid pooling (SPP). 
The CBS layer features convolution, batch normalization and activation functions. The C3 module comprises three standard convolution layers and multiple bottlenecks. 
SPP utilizes two pooling kernels, $5 \times 5$ and $1 \times 1$, which enhance the receptive field and accommodate any image aspect ratio and dimension. However, using CBS, SPP and C3 architectures in YOLOv5’s backbone can lead to increased computational complexity and memory usage, resulting in slower detection and higher resource demands. 
%Seb: Voncarlos, when you say "this paper" below do you mean "we"? If this is the case change it to this term
To address these issues and achieve a lighter network model, this paper employs ShuffleNetv2 as the backbone network. 
The neck section further refines the FPN structure to accelerate feature fusion and inference information transmission. 
The head is responsible for object detection on the feature pyramid, consisting of convolutional layers, pooling layers and fully connected layers. 
The detection head's main function is to perform multi-scale object detection on feature maps extracted from the backbone network. %In the prediction part, the Generalized Intersection over Union (GIOU) loss function is employed to assess the detection loss of the bounding box. 

\subsubsection{YOLOv8}

%Seb: below put the reference of YoloV8
YOLOv8 is the latest version of YOLO. 
It was refactored on the basis of YOLOv5, and many small strategies for improving the performance of the whole network were added. 
The latest version emphasizes the following key elements:
\begin{itemize}
    \item \textbf{Backbone}: YOLOv8 utilizes a variant of the Cross Stage Partial (CSP) DarkNet-53 network \cite{yolov88} as its backbone. 
    This architecture splits the feature map into distinct parts for convolution operations and their outputs, effectively reducing computational complexity while maintaining the detector's learning capacity.
    \item \textbf{Neck}: The neck of YOLOv8 employs the PAN-FPN module for efficient multi-scale feature fusion. 
    By combining the strengths of FPN and PAN architectures, this module allows upper layers to capture richer contextual information while lower layers preserve precise location details.
    \item \textbf{Head}: YOLOv8 features a decoupled head architecture that separates classification and detection tasks. 
    Moving away from the traditional anchor-based approach, YOLOv8 adopts an anchor-free method that detects objects based on their centers and predicts the distances to the bounding box edges, eliminating the need for predefined anchors.
\end{itemize}

Despite advancements in YOLO's architecture, newer versions still face challenges in accurately detecting animals in complex natural environments. 
Issues such as weather variations, lighting changes and animal occlusions frequently result in missed detections \cite{Ma2024}. 
To overcome this, the model needs to capture richer features. 
Recent advancements, particularly attention mechanisms, have improved performance by enhancing relevant features and suppressing noise. 
To address these limitations, we have integrated the Convolutional Block Attention Module (CBAM)~\cite{cbam}.

\subsubsection{CBAM}

\begin{figure*}
% \begin{adjustwidth}{-\extralength}{0cm}
\centering
\includegraphics[width=15cm]{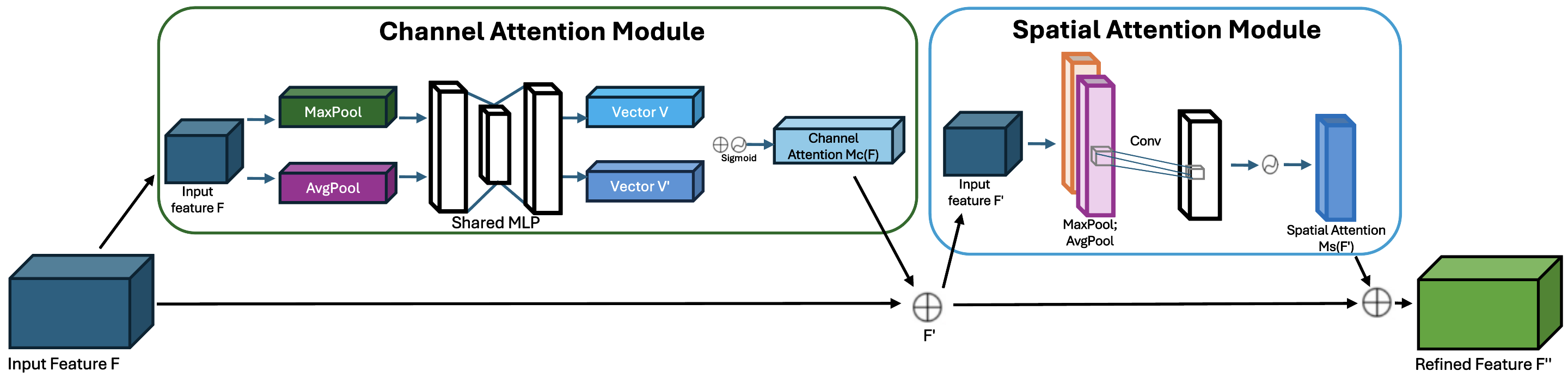}
% \end{adjustwidth}
\caption{Overview of the CBAM structure.}
\label{cbam_mecanism}
\end{figure*}

CBAM \cite{cbam} is designed to enhance the convolution layer's ability to extract important features while minimizing irrelevant information. 
To address the challenge of low significance in detecting cows due to complex backgrounds and overlapping images, CBAM introduces channel and spatial attention mechanisms after the SPP module in the YOLOv8 network model.
The CBAM architecture is illustrated in Figure \ref{cbam_mecanism}. 
This module integrates both channel and spatial attention mechanisms and serves as a lightweight add-on that can be incorporated into any CNN architecture.

The channel attention module works by pooling the input feature maps into two distinct maps using maximum and average pooling to mitigate information loss from pooling operations. 
These two feature maps are then processed by a weight-sharing Multi-Layer Perceptron (MLP) network for dimension reduction and expansion, thereby reducing the number of parameters. 
The outputs of the MLP are combined and passed through a sigmoid activation function to produce the channel attention\( M_c(F) \), which is computed as shown in Equation \ref{channelAttentionf}.

\begin{equation}
\label{channelAttentionf}
\scalemath{0.94}{ M_c(F) = \sigma \left\{ \text{MLP}\left[\text{AvgPool}(F)\right] + \text{MLP}\left[\text{MaxPool}(F)\right] \right\}}
\end{equation}

In the formula, $\text{AvgPool}(F)$ and $\text{MaxPool}(F)$ denote respectively the average pooling and maximum pooling operations applied to the feature map $F$. 
$\sigma$ represents the sigmoid activation function while $\text{MLP}()$ refers to a multilayer perceptron.
The spatial attention module focuses on emphasizing the significance of input values in the spatial dimension. 
Initially, the input values are pooled using both maximum and average operations and then concatenated. 
A 7 × 7 convolutional layer is then applied to reduce the dimensionality to a single channel. 
The final output of this module is obtained using the sigmoid activation function. 
The spatial attention feature map $M_s(F')$ is computed as shown in Equation \ref{spatialAttentionf}.
\begin{equation}
\label{spatialAttentionf}
{\scriptstyle M_s(F') = \sigma \left\{ f\left[\text{AvgPool}(F'); \text{MaxPool}(F')\right] \right\} = \sigma \left\{ f^{7 \times 7} \left[\left(F_{\text{avg}}^s; F_{\text{max}}^s\right)\right] \right\}},
\end{equation}
in which $f^{7 \times 7}()$ represents the convolution operation with a $7 \times 7$ kernel. 
$F_{\text{avg}}^s$ and $F_{\text{max}}^s$ denote the global average pooling and maximum pooling operations applied in the spatial attention mechanisms, respectively.

In conclusion, CBAM synthesizes the attention layers by combining the channel and spatial attention modules to produce the final feature map \cite{Wang2021}, which can be expressed as:
\begin{equation}
\label{f1}
F' = M_c(F) \otimes F
\end{equation}
\begin{equation}
\label{f2}
F'' = M_s(F') \otimes F'
\end{equation}

In this formula, $F'$ represents the feature map from the intermediate layer of the CNN, $\otimes$ refers to the elementwise multiplication while $F''$ denotes the feature map resulting from the CBAM output.

%To overcome the challenges of varying cow body postures and small target sizes from different camera types in the dataset, we integrated the CBAM module into the original YOLOv8 architecture, as shown in Figure \ref{Yolov8_cbam}. This integration enhances the detection of key cow features. To manage network complexity and processing speed, CBAM was added only at the end of the backbone, reducing its impact on detection speed. Consequently, the modified YOLOv8-CBAM network achieves efficient and accurate cow detection in complex environments.

Generally, a CBAM module is composed of two key components: the spatial attention and the channel attention, each playing a critical role in enhancing feature representation. 
Spatial attention works by focusing on important regions of the input image, assigning higher attention values to key spatial areas. 
This mechanism helps the model prioritize crucial parts of the object, such as a cow's distinctive features, even in complex environments with occlusions or background noise. 
In contrast, channel attention operates by assigning a weight to each feature map, enabling the model to emphasize the most relevant channels and refine the extracted features, thereby improving overall feature representation.

\begin{figure}
  \centering
  \includegraphics[width=\linewidth]{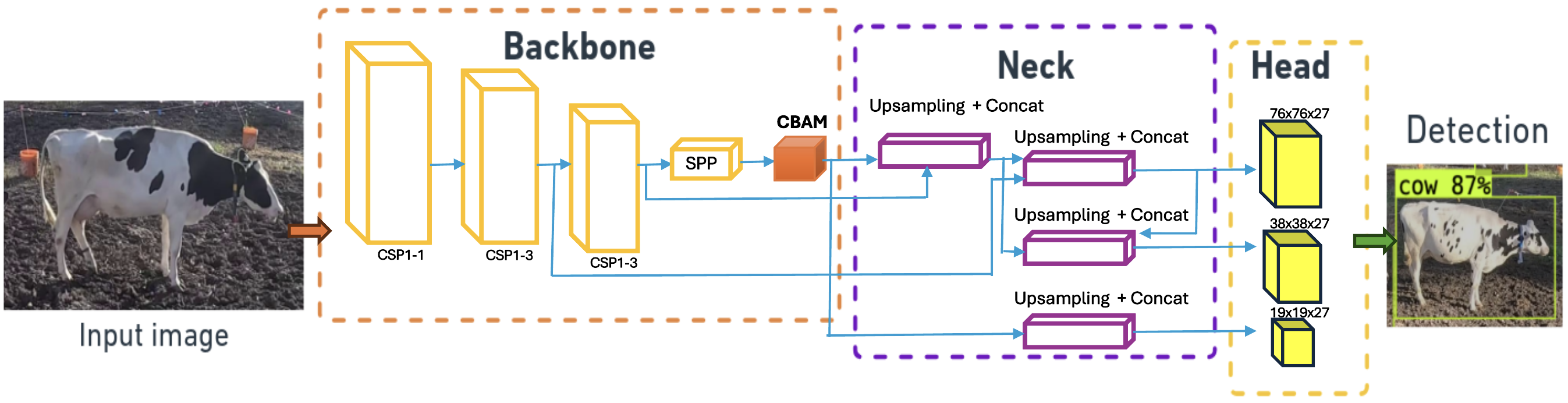}
  \caption{Tailored YOLOv8-CBAM detection model.}
  \label{Yolov8_cbam}
\end{figure}

In the specific integration of CBAM that we did with YOLOv8, this module is applied at the end of the backbone network, after the initial feature extraction layers. 
This placement, as shown in Figure~\ref{Yolov8_cbam} (orange block), ensures that the model can leverage the improved feature representation from CBAM while minimizing its impact on processing speed. 
By positioning CBAM at the final stage, the architecture maintains a relatively fast processing time, which is essential for real-time applications, such as livestock monitoring. 
This integration allows YOLOv8-CBAM to significantly enhance the cow detection accuracy, particularly in challenging, cluttered environments, while maintaining efficient performance suitable for resource-constrained deployment scenarios.

\section{Experimental Results}
\label{sect_experiments}

\paragraph{Experimental setup.}
The experiments were conducted on a dedicated Compute Canada cluster, equipped with 4 NVIDIA A100 SXM4 GPUs, each offering 40 GB of memory.
The deep learning framework utilized was based on the Ultralytics package, with PyTorch 1.11.0. 
For training, GPUs are highly recommended due to their parallel processing capabilities, which significantly accelerate the computations of the model.
The models tested, including Mask R-CNN, YOLOv5 and YOLOv8-CBAM, accept by default images with a fixed size of 640 × 640. 
This resolution helps to reduce the memory and computational load, making it feasible to train denser models on machines with more limited resources.
The Compute Canada cluster's powerful configuration was crucial for training large models and processing complex datasets efficiently. 
%Seb: we have also to add the mention to the compute Canada cluster in the acknowledgments
For inference, we used 4 GB of GPU memory, suitable for real-time edge deployment, while with respect to storage, we used 1 TB for training and 20–50 GB for inference.

To train the algorithms, two optimization methods were tested: Stochastic Gradient Descent (SGD) and Adam. 
This comparison was motivated by several findings. For example, \citet{kunstner2023noise} show that Adam, which is less dependent on the learning rate, produces more accurate gradient estimates. 
However, as noted by \citet{wilson2018margin}, Adam's generalization capabilities can be less effective compared to SGD. 
On the other hand, \citet{Mehmood2023} highlight that SGD is fast and computationally efficient, but highly sensitive to fixed learning rates. 
To address this, \citet{carvalho2020} suggest adopting a flexible learning rate that gradually decreases during training. 
A fixed set of parameters was experimentally established to ensure a fair comparison between the algorithms as shown in Table \ref{hyperparameters}. In particular, an adaptive learning rate was used, gradually decreasing from 0.01 to 0.001 during training.

In terms of the Mask R-CNN segmentation approach, ResNet-101 was used both as backbone and head architecture. 
For the YOLO version we use CSPNet as the backbone to extract features. 
In addition, pre-trained backbone weights based on the COCO data set \cite{cocodaata} were also adopted to accelerate the training process in detection models.

\begin{table}
\caption{The training parameters used in all experiments in this paper.}\label{hyperparameters}
\setlength{\tabcolsep}{30pt}

\begin{tabular}{lr}
\toprule
 Parameter & Values  \\ \midrule
 Batch size & 16  \\
 Image size & 640$\times$640  \\
 Initial learning rate & 0.01 \\
 Final learning rate & 0.001  \\
 Weight decay & 0.0005   \\
 Momentum &  0.937 \\
 
\bottomrule
\end{tabular}
\end{table}

\paragraph{Model evaluation.}
After training each model (see Figure \ref{proposedMethod} (e)), the final step is to test and validate the results (see Figure \ref{proposedMethod} (f)). 
This study evaluates model performance using precision, recall, mean average precision (mAP) and F1-score. 
These metrics are computed based on the following parameters: true positives TP (\emph{i.e.}, the cow is present in the image and detected by the model), 
%Seb: in the above when you say the cow is a specific one or simply a cow, if it is the second case we should change it to "a cow"
true negatives TN (\emph{i.e.}, the cow is neither present nor detected in the image), false positives FP (\emph{i.e.}, the cow is not present in the image but wrongly detected by the model), and false negatives FN (\emph{i.e.}, the cow is present in the image but not detected by the model).

\emph{Precision} measures the proportion of correctly predicted bounding boxes for cow objects out of the total number of predicted bounding boxes in the cow datasets.
\begin{equation} 
Precision = \frac{TP}{TP+FP}.
\end{equation}

\emph{Recall} indicates the proportion of correctly predicted cow bounding boxes out of all actual cow instances in the dataset.
\begin{equation} 
Recall = \frac{TP}{TP+FN}.
\end{equation}
The \emph{F1-score} is the harmonic mean of precision and recall, providing a single metric that balances both measures in the evaluation of the cow detection model.
\begin{equation} 
F_1 = \frac{ 2 \times 
 \text{Precision} \times \text{Recall}}{\text{Precision} + \text{Recall}}.
\end{equation}
\emph{Mean Average Precision} (mAP) assesses the cow detection model by averaging precision across different recall levels. 
It is measured at an intersection over union (IoU) threshold of 0.50 (mAP@0.50) or across a range of IoU thresholds from 0.5 to 0.95 (mAP@0.5:0.95).
\begin{equation} 
\text{mAP} = \frac{\sum_{i=1}^{C} \text{AP}_i}{C} .
\end{equation}
Here, $AP$ represents the average precision for the $i$-th cow category, and $C$ denotes the total number of cow categories.

\subsection{Comparison One-Stage and Two-Stage Detectors}

To validate the effectiveness of the proposed approach, a comparison was conducted between Mask R-CNN, YOLOv5 and YOLOv8 detectors. 
We chose YOLOv5n and YOLOv8n for its nano-compact size, recognizing that while it offers high performance for object detection, it does come with a trade-off in accuracy compared to the larger models in the YOLO series, such as YOLOm (medium), YOLOl (large) and YOLOx (extra). 
To ensure a fair comparison, all models were trained on the cow dataset with 250 epochs and validated on raw and augmented data. 
The evaluation aimed to determine which detector, the one-stage YOLO or the two-stage Mask R-CNN, possesses greater representational capability. 
The results, presented in terms of mAP, precision, recall and inference time, are shown in Table \ref{tab3}.

\begin{table*}
\caption{Comparative results between Mask R-CNN, YOLOv8n and  YOLOv5n detection models on raw and augmented data.\label{tab3}}

\begin{tabularx}{\textwidth}{lCCCCCC}
\toprule
	\textbf{Model} &\textbf{Data type} &\textbf{mAP@0.5}	&\textbf{mAP@0.5:0.95}	& \textbf{Precision (\%)} & \textbf{Recall (\%)} & \textbf{Time (ms)}	\\
\midrule
\multirow{2}{*}{Mask R-CNN}	& Raw & 77.66 & 68.22	& 79.67		 & 80.33	& \multirow{2}{*}{4.4}					\\ 
	& Augmented & 84.89 & 73.13	& 82.33		 & 81.41	& 			\\ 
\midrule
\multirow{2}{*}{YOLOv5n}	& Raw & 85.2 & 74.4	& 83.1		 & 82.0 & \multirow{2}{*}{	1.8	}		\\
& Augmented 	 & 94.5 & 77.6	& 89.7	 & 89.1 & \\
\midrule

\multirow{2}{*}{YOLOv8n}	& Raw & 89.8 & 71.1	& 82.6		 & 86.7 & \multirow{2}{*}{1.5}				\\
	& Augmented & 94.7 & 78.1	& 92.2		 & 89.6 & \\
\bottomrule
\end{tabularx}
\end{table*}
\unskip

The results show that one-stage YOLO models significantly outperform Mask R-CNN on the cow dataset, including both raw and augmented data types. 
While data augmentation technique improves Mask R-CNN’s accuracy to 84.89\%, its mAP@0.5 on raw data decreases by 7.2\%. 
Mask R-CNN’s reliance on region proposals and mask generation often results in difficulties with occlusions and background clutter, leading to less efficient detection in challenging environments. 
Conversely, YOLO’s use of anchor boxes and grid-based predictions allows it to better handle diverse cow sizes and complex backgrounds, achieving superior results even with raw data with mAPs@0.5 of 89.8\% for YOLOv8 and 85.2\% for YOLOv5.

The application of data augmentation led to substantial improvements across all models by generating diverse and synthetic samples, thereby enhancing the models' resilience to variations in perspectives and angles. 
This ensured consistent performance across various camera sets, in indoor or outdoor environments. 
Although Mask R-CNN produced comparable results, its mAPs@0.5 are up to 10\% lower compared to the augmented data type, and its inference time is nearly twice as long as that of YOLOv5 and YOLOv8. 

\begin{figure*}
  \centering
  \includegraphics[width=\linewidth]{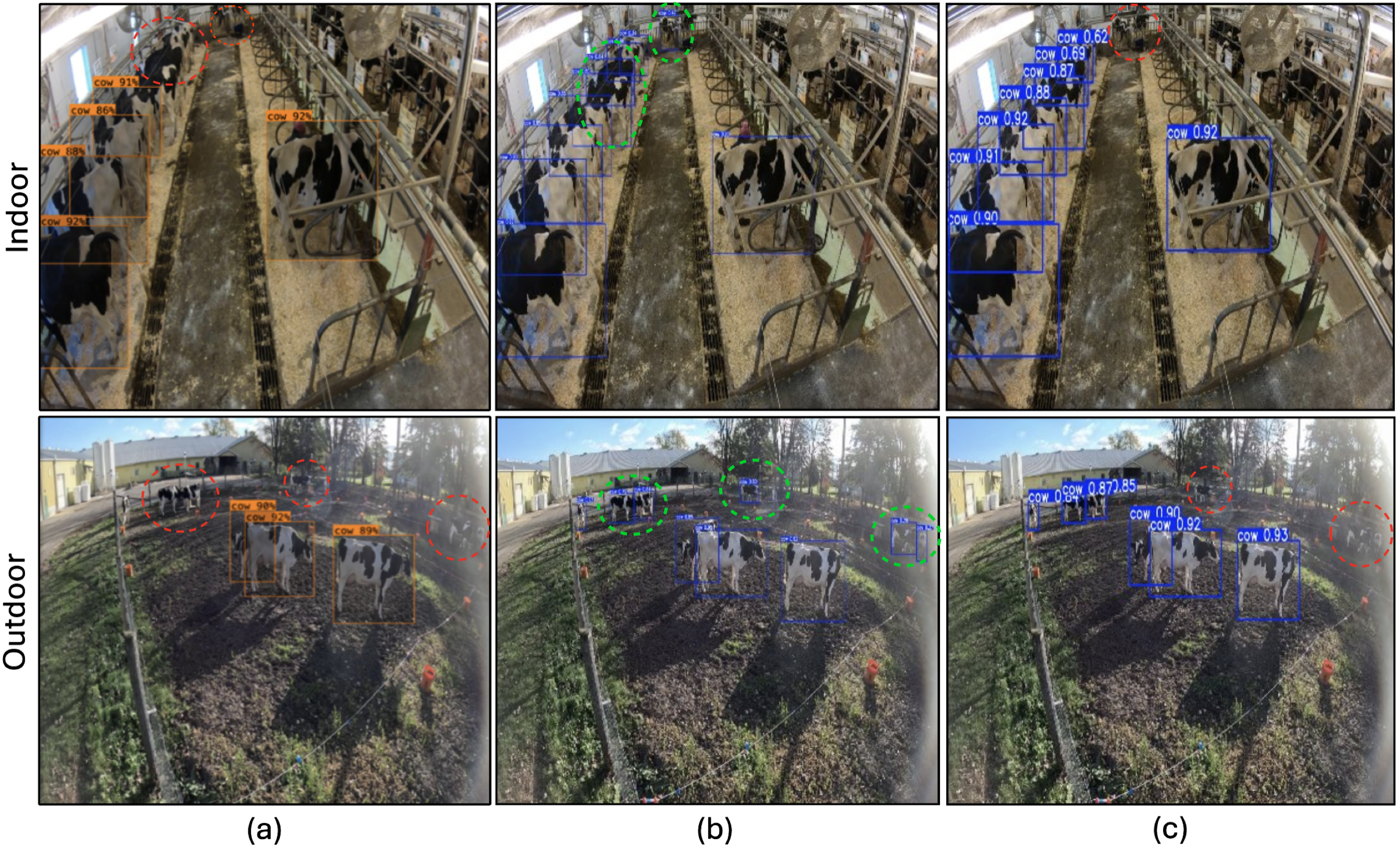}
  \caption{Visualization map using augmented data: two-stage detector:(a)  Mask R-CNN; one-stage detector: (b) YOLOv8n and (c) YOLOv5n. Missed detections are marked with red dotted circles, while correctly detected cows are highlighted with green dotted circles.}
  \label{maskrcnnPREDS}
\end{figure*}

This conclusion is reinforced by the detection predictions made in indoors and outdoors scenarios by Mask R-CNN, illustrated in Figure~\ref{maskrcnnPREDS} (a). 
Mask R-CNN may struggle with detecting small, distant objects due to its multi-stage process and feature map resolution limitations. 
Indeed, the RPN can struggle to generate accurate proposals for small objects, and pooling operations may lose essential details, complicating accurate segmentation. 
In contrast, YOLOv8 (Figure~\ref{maskrcnnPREDS} (b)) and YOLOv5 (Figure~\ref{maskrcnnPREDS} (c)) excels in detecting small objects due to its one-stage design, which directly predicts object locations and classifications and allows to handle a broader range of object scales more effectively while preserving detail across different scales.

In Table \ref{tab3}, YOLOv8 outperforms YOLOv5 in cow dataset due to its enhanced feature pyramid network (FPN), which provides superior multi-scale feature representation, making it more effective at detecting animals of varying sizes. 
Furthermore, its advanced post-processing techniques, such as enhanced non-maximum suppression (NMS) and improved handling of overlapping objects, reduce false positives and increase the accuracy of identifying multiple animals within a single image. 
This improvement is evident when comparing outdoor images in Figures~\ref{maskrcnnPREDS} (b) and (c), in which YOLOv8 effectively detects overlapping cows in low lighting conditions. 
We highlight all missed detections with red dotted circles in Figure~\ref{maskrcnnPREDS} for model comparison. 
In conclusion, YOLO architectures exhibit greater generalization compared to Mask R-CNN. 
Their superior performance, achieved by adapting to diverse conditions and effectively distinguishing objects in complex backgrounds, makes them the recommended primary architecture for this work.

\subsection{Comparison of the SGD and Adam optimizers}

In this section, the performance of both optimizers is compared across all density variations provided by the YOLO versions, ranging from the sparsest model, YOLOn (Nano), to the densest, YOLOx (Extra Large), for both YOLOv5 and YOLOv8 detectors. 
Note that this evaluation was conducted on the augmented dataset, which was selected to assess in fair manner the effectiveness of the algorithms being compared. 
The results, shown in Table~\ref{tableSGDxADAM}, highlight that using the SGD optimizer yields better outcomes. 
Notably, the best-performing model is YOLOv8x, which demonstrates the best detection precision, achieving 96.0\% mAP@0.5 and 81.0\% mAP@0.95. This suggests that SGD outperforms Adam.

In the complex cow detection scenario, the SGD's inherent regularization prevents overfitting to training data, thus improving generalization to unseen scenarios. 
Additionally, the SGD's stochastic nature allows it to escape local minima in the loss function landscape more effectively than Adam, which may become trapped due to its adaptive learning rates. 
While Adam is known for its fast convergence, it can lead to overfitting and may overlook fine-grained details critical for small object detection. In contrast SGD, as slower but more stable learning dynamics, which improves generalization by thoroughly exploring the loss landscape and capturing subtle features contributing to more effective training.

We also want to emphasize the impact of th architecture size, as we observed an improvement in detection accuracy when moving from the shallowest to the deepest models—by nearly 2.8\% from YOLOv8s to YOLOv8x and 1.7\% from YOLOv5s to YOLOv5x in mAP@0.95. 
Our choice of YOLOv8l is directly tied to its ability to strike a balance between implementation speed in real-world applications and detection precision, which in turn enhances the model's generalization capability.

%\subsection{Embedding the EMA attention module}

\subsection{YOLOv8 Fused with CBAM}

\begin{figure}
  \centering
  \includegraphics[width=\linewidth]{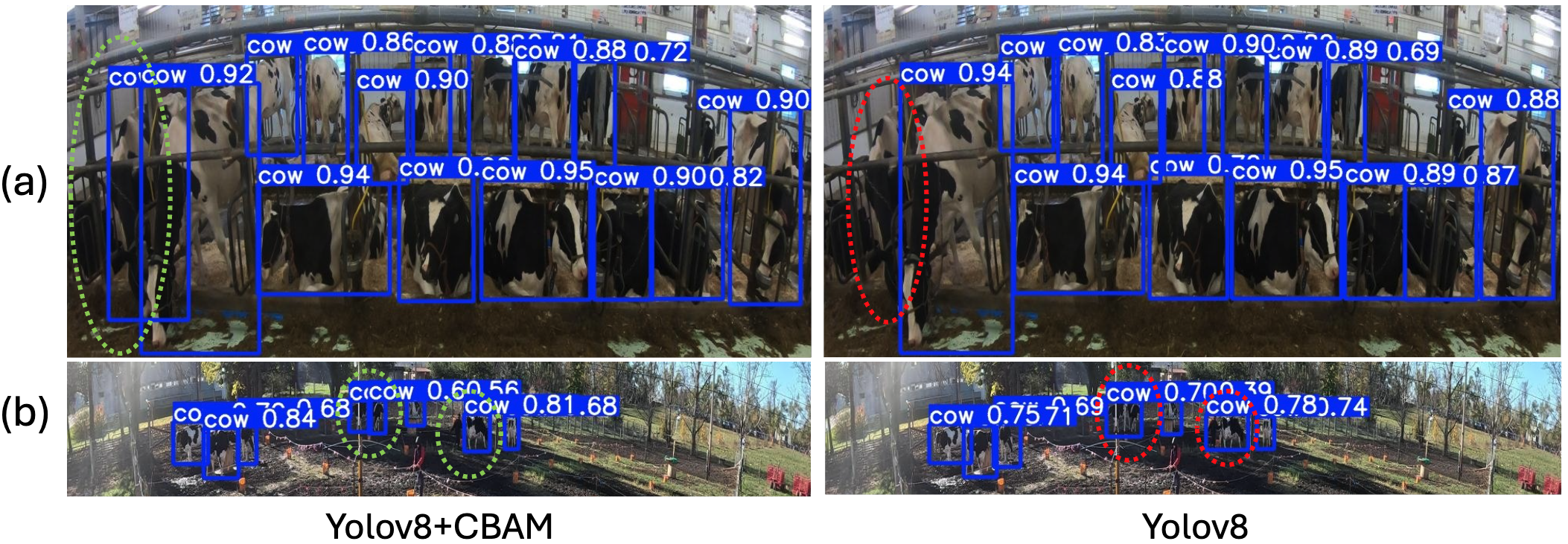}
  \caption{Illustration of Yolov8l-CBAM versus Yolov8l.}
  \label{comparisonCBAM}
\end{figure}

To evaluate the effectiveness of the CBAM attention module, we compare the performance of the best model, YOLOv8l, with and without the CBAM module.
The proposed model, YOLOv8l-CBAM, integrates the CBAM attention module into the YOLOv8l backbone. 
The comparison results, along with other baseline detection models, are summarized in Table \ref{finaltab}. 
YOLOv8l with CBAM shows improvements across all metrics. 
This enhancement is attributed to CBAM’s focus on deeper features, which addresses the inaccuracies observed with YOLOv8l when the attention mechanism is not applied. 
Additionally, Figure \ref{comparisonCBAM} provides visual comparisons of detections in complex indoor and outdoor scenarios, respectively, illustrating the results of applying CBAM with YOLOv8l compared to the baseline model YOLOv8l. 
Additionnally, Figure \ref{comparisonCBAM} (a) illustrates various occlusions, lighting conditions and pose orientation issues, which are effectively handled by YOLOv8l-CBAM. 
In Figure \ref{comparisonCBAM} (b), challenges related to scale and camera angles are present. 
However, YOLOv8l-CBAM successfully detects cows even when they are very close and obscured by people, as indicated by the green dotted circle. 
Our proposed YOLOv8l-CBAM model achieves a precision of 95.2\% and an mAP@0.5:0.95 of 82.6\%, which are respectively improvements of 2.1\% and 1.8\%, over the original YOLOv8l model.

The inference time, measured in milliseconds (ms), refers to the time taken by a model to process a single input image and generate predictions.
The inclusion of the CBAM module in YOLOv8+CBAM slightly increases the computational complexity due to the added attention mechanism. 
CBAM enhances feature representation by focusing on key spatial regions and channels within the image, improving the model's ability to manage occlusions and complex backgrounds. 
As a result, YOLOv8l + CBAM shows a marginally higher inference time of 1.7 ms compared to YOLOv8l at 1.5 ms. 
However, YOLOv8l + CBAM still outperforms previous models like YOLOv5l (1.8 ms) and Mask R-CNN (4.4 ms), achieving a precision of 95.2\%, a mAP@0.5 of 96.8\% and a mAP@0.5:0.95 of 82.6\%. 
While this higher complexity may impact performance on devices with limited resources, it still leads to low inference times, with YOLOv8 and YOLOv8+CBAM models achieving respectively fast processing speeds of 1.5 ms and 1.7 ms,enabling real-time livestock monitoring. 
Additionally, CBAM significantly enhances detection accuracy, particularly in challenging scenarios

\begin{table*}
\caption{Comparative results between detection models. \label{finaltab}}

\begin{tabularx}{\textwidth}{lCCCCC}
\toprule
	\textbf{Model} &\textbf{mAP@0.5}	&\textbf{mAP@0.5:0.95}	& \textbf{Precision (\%)} & \textbf{Recall (\%)} & \textbf{Time (ms)}	\\
\midrule
Mask R-CNN	 & 84.8 & 73.1	& 82.3		 & 81.4	& 4.4			\\

YOLOv5l	 & 95.4 & 80.3 	& 91.0	 & 87.2	& 1.8			\\

YOLOv8l	 & 96.0 & 80.8	& 93.1	 & 91.6	& 1.5			\\

YOLOv8l + CBAM	 & 96.8 & 82.6	& 95.2	 & 92.7	& 1.7			\\

\bottomrule
\end{tabularx}
\end{table*}
\unskip

\subsection{Robustness in Indoor and Outdoor Cameras}
\label{indorandoutdoors}

The final comparison in this study evaluates the performance of our approach in generability detection across different camera types, considering both indoor and outdoor environments. 
Table \ref{tab:results} summarizes the outcomes for the two most effective one-stage detectors (YOLOv8l-CBAM and YOLOv8l), measured in terms of precision, recall and F1-score. 
For these experiments, we have trained a single model capable of detecting cows across various camera scenarios, both indoors (IW, IP and IC) and outdoors (OP, OE and OC), which ensures the model's generalization and robustness across six different camera types. The different cameras used in the experiment include Interior Wall (IW), Interior Pipe (IP), Interior Ceiling (IC), Outdoor Pipe (OP), Outdoor Enrichment (OE), and Outdoor Corridor (OC). These acronyms will be consistently used throughout the text for brevity.

\begin{figure*}
  \centering
  \includegraphics[width=\linewidth]{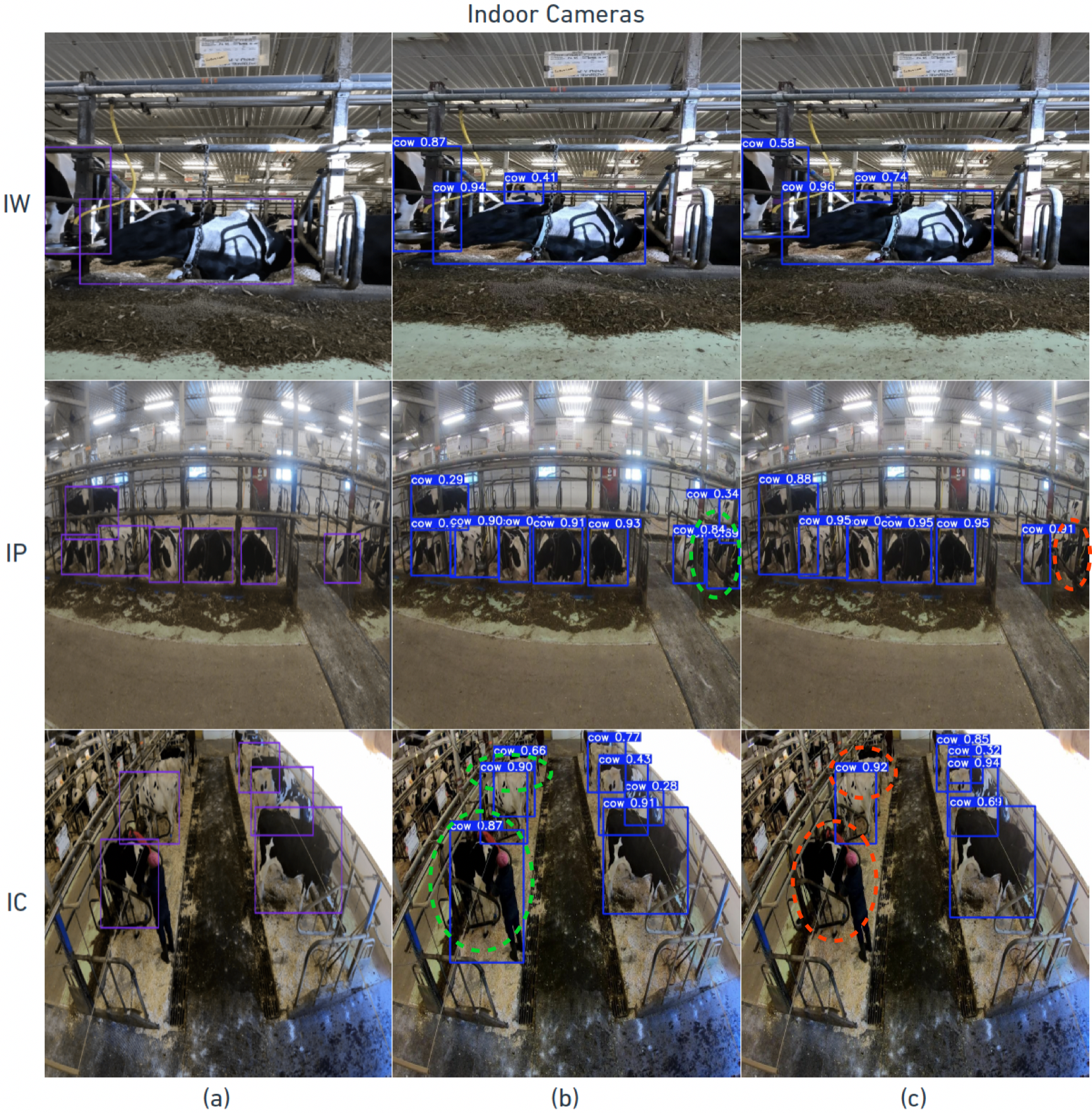}
  \caption{Indoor cameras: (a) raw image with the ground-truth; (b) YOLOv8l-CBAM detection prediction and (c) YOLOv8l detection prediction. 
  Missed detections are marked with red dotted circles, while correctly detected cows are highlighted with green dotted circles.}
  \label{indoorC}
\end{figure*}

To further validate the effectiveness of our model in detecting cows, we selected images showcasing cows in various scenarios and behaviors. 
We performed a comparative analysis between YOLOv8l-CBAM and YOLOv8l, using visualizations to provide a clearer model comparison and visual explanation. 
First, the detection outcomes are depicted to indoors cameras (IW, IP and IC) as shows in Figure \ref{indoorC}, in which Figure \ref{indoorC} (a) represents the raw image with the ground-truth annotations in purple color, Figure \ref{indoorC} (b) illustrates the detection results of YOLOv8l-CBAM, while Figure \ref{indoorC} (c) showcases the detection results of YOLOv8l.

\begin{figure}
  \centering
  \includegraphics[width=\linewidth]{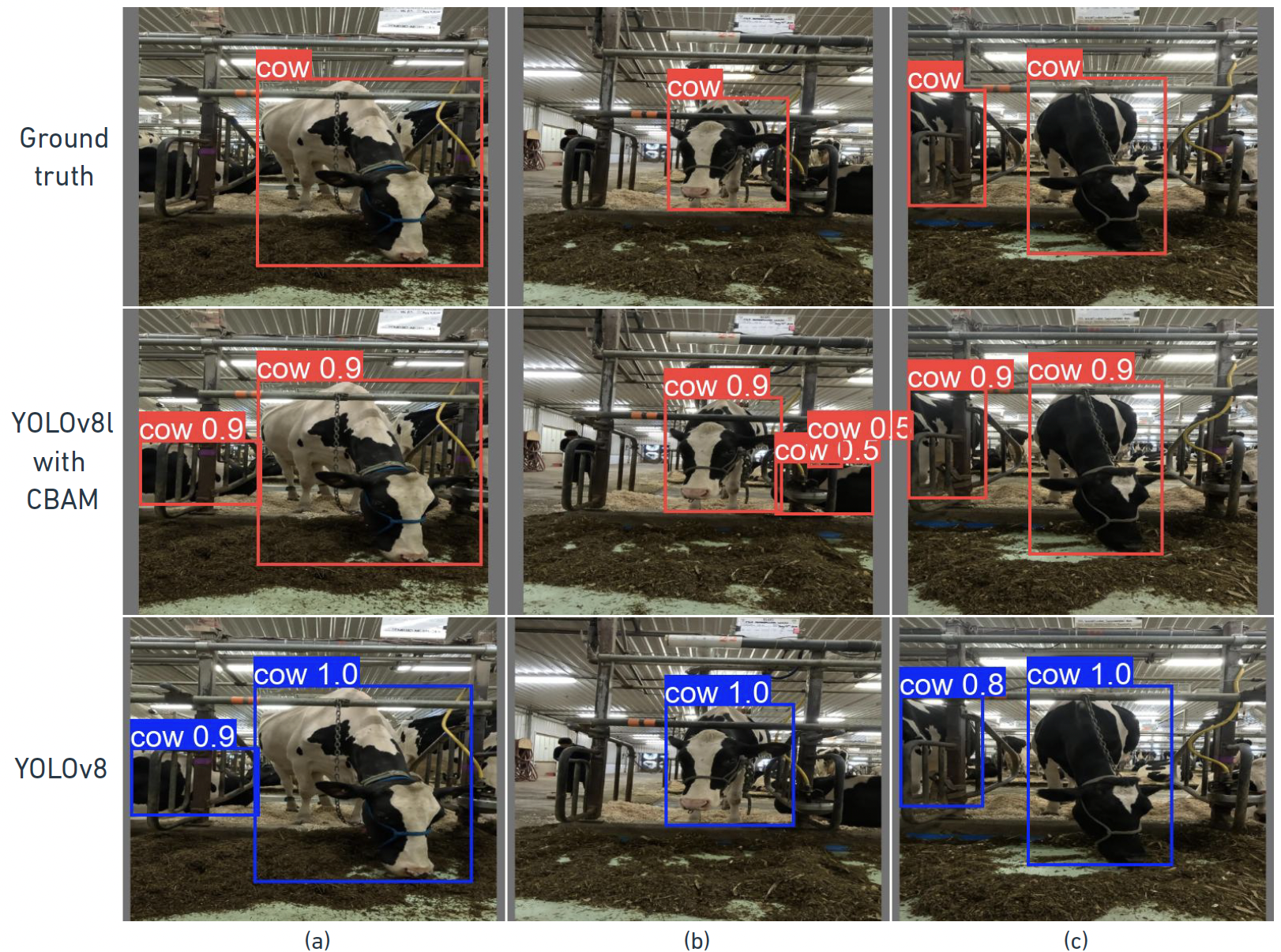}
  \caption{Detection comparison: YOLOv8 vs. YOLOv8l+CBAM. (a) Sample 1, (b) Sample 2 and (c) Sample 3.}
  \label{falsepositiveIW}
\end{figure}

%Seb: as mentioned previously the acronym for the different cameras should be detailed the first time they are used
%voncarlos: I added: The different cameras used in the experiment include Interior Wall (IW), Interior Pipe (IP), Interior Ceiling (IC), Outdoor Pipe (OP), Outdoor Enrichment (OE), and Outdoor Corridor (OC). These acronyms will be consistently used throughout the text for brevity.
In the case of the interior wall (IW) camera, individual cows are present but often appear under poor lighting conditions, with occluded parts and other cows frequently visible in the background. 
These factors present challenges for the YOLO detector models.
Unexpectedly, the IW camera resulted in several detections of cows that had not been manually identified by the inspector (ground truth), resulting in an increase in false positives as shown on Figure \ref{falsepositiveIW}.
This accounts for the lower precision scores for this camera, with YOLOv8l-CBAM and YOLOv8 achieving respectively 89.6 and 88.2 in precision (P\%), as shown in Table \ref{tab:results}.

\begin{table*}
\caption{Summary of results for all camera types.}
\centering
\renewcommand{\arraystretch}{1.4}
\begin{tabular}{|c|p{0.4cm}|p{0.4cm}|p{0.5cm}|p{0.4cm}|p{0.4cm}|p{0.5cm}|p{0.4cm}|p{0.4cm}|p{0.5cm}|p{0.4cm}|p{0.4cm}|p{0.5cm}|p{0.4cm}|p{0.4cm}|p{0.5cm}|p{0.4cm}|p{0.4cm}|p{0.5cm}|}
\hline
\textbf{Camera Type} & \multicolumn{3}{c|}{\textbf{IW}} & \multicolumn{3}{c|}{\textbf{IP}} & \multicolumn{3}{c|}{\textbf{IC}} & \multicolumn{3}{c|}{\textbf{OP}} & \multicolumn{3}{c|}{\textbf{OE}} & \multicolumn{3}{c|}{\textbf{OC}} \\ \cline{1-19} 
                                      & \textbf{P\%}  & \textbf{R\%}  & \textbf{F1\%} &  \textbf{P\%}  & \textbf{R\%}  & \textbf{F1\%} & \textbf{P\%}  & \textbf{R\%}  & \textbf{F1\%}  & \textbf{P\%}  & \textbf{R\%}  & \textbf{F1\%} & \textbf{P\%}  & \textbf{R\%}  & \textbf{F1\%} & \textbf{P\%}  & \textbf{R\%}  & \textbf{F1\%}  \\ \hline
YOLOv8l-CBAM & 89.6  & 89.0  & 89.2 & 93.5  & 91.2 & 92.3 & 85.3  & 82.9  & 84.9 & 88.2 & 96.2 & 92.0  & 85.7  & 99.8  & 99.9  & 88.1  & 92.0  & 90.0    \\  \hline

YOLOv8l & 88.2  & 90.5  & 89.3 & 93.5  & 88.3  & 90.8 & 83.8  & 81.5  & 82.6 & 91.0 & 91.1  & 91.0  & 100  & 84.4  & 91.2  & 88.0  & 92.8 & 90.3   \\  \hline
\end{tabular}

\label{tab:results}
\end{table*}

For the interior pipe (IP) camera, many individual cows are showed in different poses and orientations.
Overall, the precision, recall and F1-score reach up to 90\% for both YOLOv8l-CBAM and YOLOv8l, which can be considered to be good results considering the total number of training images.
This IP camera set, producing 529 augmented images, is one of the largest among the camera types.
This allows for enhanced model capability, leading to improved detection performance in this scenario. 
In the ground-truth (Figure \ref{indoorC} (a)), 7 cows were identified, but the YOLOv8l-CBAM model detection correctly predicted 9 cows (Figure \ref{indoorC} (b)). 
However, YOLOv8l mistakenly merged 2 cows on the left side into a single cow, resulting in an error, and it also failed to detect 2 cows in the corner on the right side (Figure \ref{indoorC} (c)). 
Possibly, this can occur because YOLOv8l might have difficulty generalizing in scenarios with significant occlusions due to its lack of an attention mechanism module.

The last indoor camera, interior ceiling (IC), presents the most challenging scenario due to the complex background and the camera’s angle, which captures the sides of cows in a crowded environment where they are positioned close to one another. 
This is reflected in the results demonstrated in Table \ref{tab:results}, with respectively F1-scores of 84.9 for YOLOv8l-CBAM and 82.6 for YOLOv8l.
As shown in Figure \ref{indoorC} (c), YOLOv8l struggles to detect when a person is occluded by a cow and exhibits low detection confidence. 
These lower results for this specific camera setup can be attributed to the limited number of training samples available as detailed in Table \ref{tabledataset}. 
However, YOLOv8l-CBAM shows improvements respectively of 1.5\% in precision, 1.4\% in recall and 2.3\% in the F1-score, for the same cow dataset.

\begin{figure*}
  \centering
  \includegraphics[width=\linewidth]{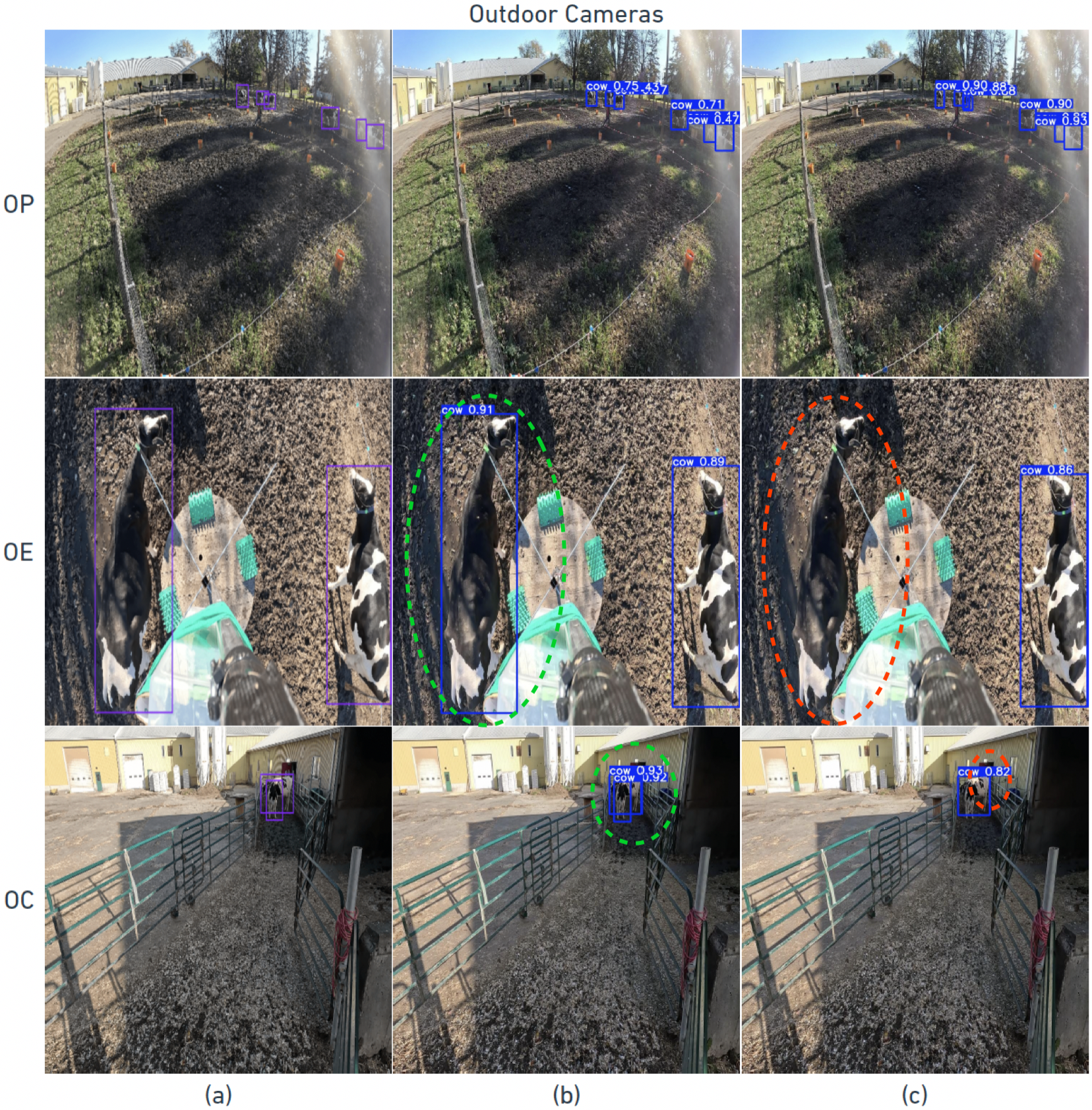}
  \caption{Outdoor cameras: (a) raw image with the ground-truth; (b) YOLOv8l-CBAM detection prediction and (c) YOLOv8l detection prediction.
  Missed detections are marked with red dotted circles, while correctly detected cows are highlighted with green dotted circles.}
  \label{outdoorC}
\end{figure*}

Outdoor cameras present unique challenges, including variations in scale and size, occluded objects, lighting conditions and different camera angles and heights. 
In Table \ref{tab:results}, all outdoor cameras achieve F1-scores of up to 90\%, indicating that the model performs well in terms of both precision and recall across these conditions. 
For the outside paddock (OP) camera, as shown in first line of Figure \ref{outdoorC}, many cows are distant from the camera, with occluded objects and glare in the image corners. 
Despite these challenges, both detection models performed accurately in these situations.

Surprisingly, the results for the outside enrichment (OE) camera indicate that the model performs well in detecting cows from a top-down perspective, which positively impacts the model's generalization.
Figure \ref{outdoorC} (b) shows two cows from different perspectives: one black cow in a shadowed area and another black-and-white cow with illumination. 
In both cases, the detection was accurate using YOLOv8l-CBAM.

Finally, the outside corridor (OC) camera is one of the most challenging on outdoor scenarios due to varying lighting conditions, cows being far from the camera, and objects like wire fences that can significantly obstruct detection. 
Despite these difficulties, the results shown in Table \ref{tabledataset} demonstrated precision scores of 88.1\% for YOLOv8l-CBAM and 88.0\% for YOLOv8l. 
As illustrated in Figure \ref{outdoorC} (c), YOLOv8l struggles with occlusions in shadowed areas and small objects at a distance. 
However, overall, YOLOv8l performs well in complex scenarios and shows similar accuracy to YOLOv8l-CBAM.

\subsection{Comparative Analysis of Model Performance}
\label{comparisonMethods}

\begin{figure*}
  \centering
  \includegraphics[width=\linewidth]{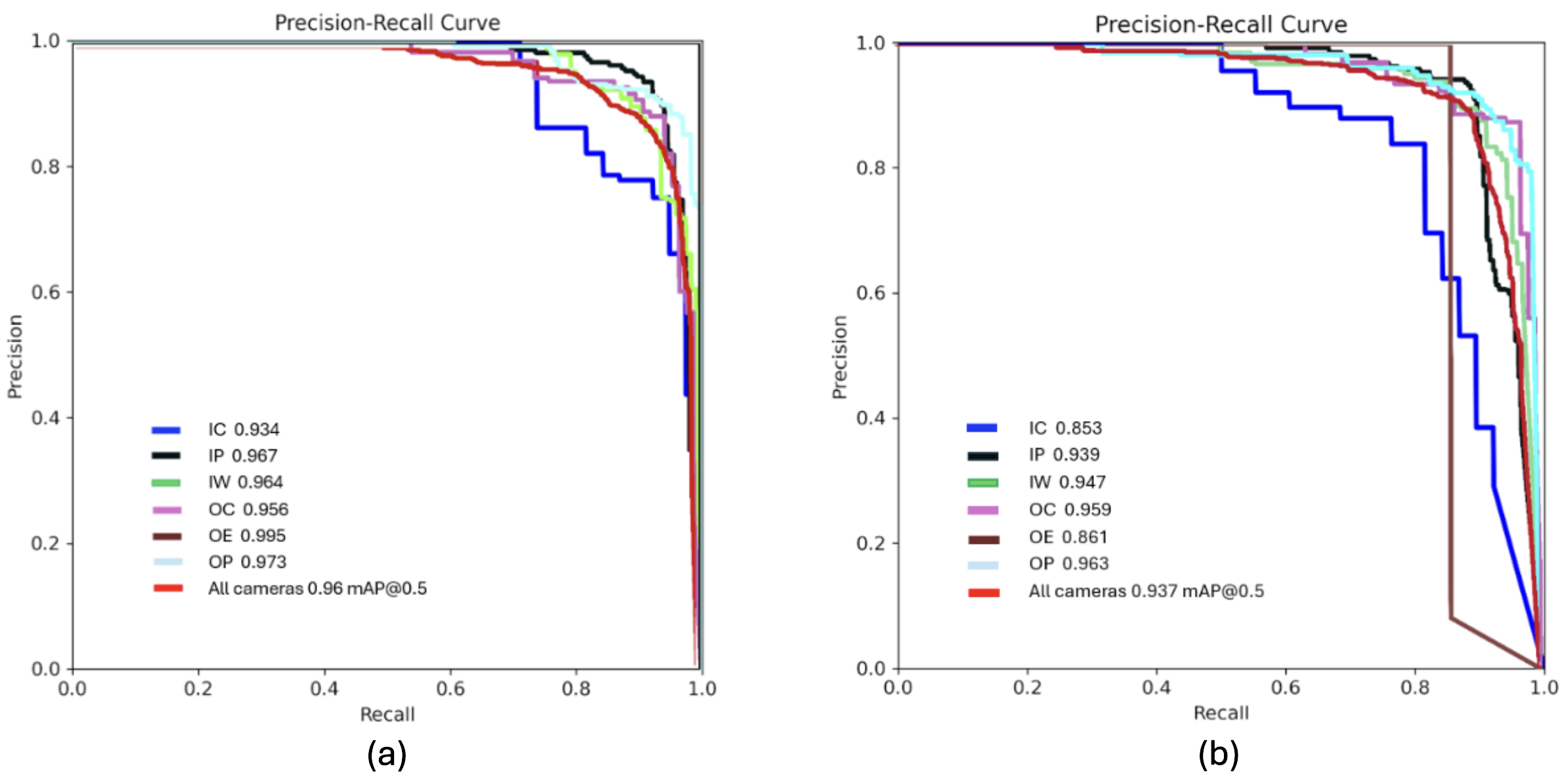}
  \caption{(a) P–R curve of YOLOv8l-CBAM; (b) P–R curve of YOLOv8l.}
  \label{COMPARISONmap05}
\end{figure*}

The precision–recall (P–R) curves for YOLOv8l-CBAM and YOLOv8l were plotted with precision and recall respectively on the x-axis and y-axis, as shown in Figure \ref{COMPARISONmap05} in which the graphics illustrate varied improvements in AP across different cameras. 
%Seb: the table should be reduced a little bit in size to avoid it going over the second column
%Voncarlos: ok
\small
\begin{table}[width=.9\linewidth,cols=3,pos=h]
\caption{Evaluation results of various optimizers on YOLO architecture densities. 
}\label{tableSGDxADAM}
\begin{tabular}{@{}l p{0.9cm} p{0.9cm} c p{0.9cm} p{0.9cm} p{0.9cm}@{}}
\toprule
  & \multicolumn{3}{c}{SGD} & \multicolumn{2}{c}{Adam} & Time \\
\cline{2-3} \cline{5-6}
Models & mAP0.5 & mAP0.95 & & mAP0.5 & mAP0.95 & (ms) \\
\midrule
YOLOv8n   & 94.7 & 78.1 & & 93.1 & 77.2 & 1.2 \\
YOLOv8m   & 96.1 & 79.6 & & 94.8 & 77.6 & 1.2 \\
YOLOv8l   & 96.0 & 80.8 & & 95.1 & 79.1 & 1.5 \\
YOLOv8x   & 96.0 & 81.0 & & 94.4 & 76.1 & 1.9 \\
\midrule
YOLOv5n   & 94.5 & 77.6 & & 93.1 & 75.2 & 1.7 \\
YOLOv5m   & 95.0 & 80.3 & & 94.2 & 79.1 & 1.7 \\
YOLOv5l   & 95.4 & 80.3 & & 93.1 & 77.7 & 1.8 \\
YOLOv5x   & 95.6 & 79.3 & & 93.1 & 76.6 & 2.2 \\
\bottomrule
\end{tabular}
\end{table}

The area under the P–R curve indicates the AP, in which a larger area signifies higher AP values. 
Notable camera detections with minor changes in AP for both models include IW, OC and OP, indicating only slight differences in detection predictions for these cameras. 
However, IC and OE cameras yielded results below 90\% in YOLOv8l, likely due to the complexities discussed in Section \ref{indorandoutdoors}. 
For instance, the OE camera achieved a peak AP of 99\% with the YOLOv8l-CBAM model, showing only a marginal improvement. 
Significant AP differences were observed in indoor cameras: IC, IP and IW, with notable increases of respectively 8.1\%, 2.8\% and 1.7\%, when using YOLOv8l-CBAM.

The results for the IC camera are below expectations due to the imbalanced dataset used in the analysis. 
The IC camera was trained with a limited number of samples, specifically only 66 augmented images in the training set, which contributes to the poor performance observed in the precision-recall (P-R) curve, as represented by the blue line. 
In contrast, the improvements on YOLOv8l-CBAM can be attributed to the optimized architecture. 
In addition, YOLOv8l-CBAM enhances interpretability by showing which input parts are most influential through attention weights and effective regularization methods help it avoid overfitting and perform better when fewer samples are available. 
However, YOLOv8 advanced architecture may require larger datasets to achieve its full potential, making it less suited for scenarios with limited data. 
In summary, the proposed YOLOv8l-CBAM outperforms YOLOv8l by 2.3\% in overall mAP across all camera types, demonstrating its effectiveness in enhancing cow detection accuracy and its superior ability to generalize across different camera variations.

\section{Discussion}
\label{sect_discussion}

\subsection{Limitations and Potential Improvements}

Automatic animal detection reduces the need for human resources and specialized expertise in feature analysis. 
Due to the advance of monitoring technologies for welfare dairy cows, such as trap cameras and long-range drones, the availability of richer and more diverse image data is increasing. 
This enhancement in data quality is anticipated to improve the performance of detection models over time.

At this stage, our task of detecting welfare dairy cows relies solely on features extracted from cow image data. 
However, there is a wealth of additional expert knowledge that could enhance cow detection and classification \cite{neethirajan2024artificial}. 
%Seb: Voncarlos please include a few references of papers in which these types of data are used for a similar task
%voncarlos:ok
This includes relying on sensors and metadata, and integrating data such as GPS, RFID or thermal imaging, without causing discomfort to the animals \cite{rohan2024application}. 
Compared to merely improving the model structure, incorporating this expert knowledge could significantly boost performance. 
However, the current model prioritizes maintaining cow welfare and does not allow for the use of invasive methods to aid detection.

Regarding the indoor ceiling (IC) camera, the limited training data for this specific camera type posed challenges, particularly in crowded and occluded scenarios. 
To mitigate this, we employed data augmentation and transfer learning techniques. 
While data augmentation enhanced overall performance, as evidenced in Table \ref{tab3}, the small dataset constrained substantial improvements.
Transfer learning, utilizing pre-trained models, partially addressed the unique challenges associated with the IC camera. 
Nevertheless, a larger and more balanced dataset is essential to improving detection accuracy for this camera type, underscoring the importance of addressing data imbalance in such scenarios.

To further address these challenges, lightweighting the model by optimizing its architecture and reducing computational demands is a key direction for improvement. 
%Seb: put the reference for the MobileNet architecture
%voncarlos:ok
For example, leveraging efficient architectures such as MobileNet \cite{howard2017mobilenetsefficientconvolutionalneural} or pruning techniques could reduce resource consumption while maintaining accuracy. 
Additionally, as mentioned incorporating multimodal data or thermal imaging could enhance model operability in real-world applications, as the fusion of image data with these modalities would not only improve detection robustness but also provide insights into cow behavior and health. 
Future work could explore integrating multimodal frameworks into current models to increase precision and utility in complex farm environments.

While we have made progress in model lightweighting by reducing the number of parameters and computational demands, more effective strategies for enhancing computational efficiency and accuracy remain, such as using hybrid models and ensemble techniques are possible. 
These approaches could integrate YOLO with other detectors, leveraging each model’s strengths to improve overall performance in complex or dense detection tasks. 
%Seb: Voncarlos can you put a reference to these experimental results?
%voncarlos:ok
Experimental results in Section \ref{indorandoutdoors} shows that the current model excels with small or partially occluded targets, demonstrating faster convergence and better accuracy in these scenarios. 
While performance may decrease with limited training data, attention mechanisms can improve this effect as well as performance in particular contexts. 
Consequently, additional enhancements are required in future iterations.

In practical livestock farming, detecting key anatomical regions of cows is crucial for monitoring their health. 
For example, lightweighting combined with multimodal integration could be especially beneficial for detecting critical areas such as legs (to monitor hoof issues affecting milk production \cite{impact19}) and heads (to track rumination behavior \cite{impact21} and identify individual cows \cite{impact22,impact23}). T
his approach could also provide actionable insights into cow behavior patterns and facilitate better livestock management, even in challenging and imbalanced scenarios.

\subsection{Future Research Directions} 

AI-based detection techniques that are noninvasive minimize the need for human intervention, allowing for efficient and accurate data collection, thereby setting a new standard in enhancing animal welfare \cite{Zemanova2020}. 
However, the use of AI techniques also presents certain limitations. 
As demonstrated in this study, computer vision methods like CNNs can effectively detect animals but may struggle in complex environments \cite{Neethirajan2021,Siriani2022}. 
Developing and training AI systems requires substantial amounts of data \cite{upadhyay2019artificial}. 
However, data can be scarce or difficult to obtain, which poses significant challenges to the effectiveness of AI applications \cite{Sarker2022}.

To address the challenges of limited and imbalanced data, future research should focus on advanced data augmentation techniques, synthetic data generation and transfer learning to optimize performance in data-scarce scenarios. 
Leveraging synthetic datasets created through generative models or simulation environments can supplement real-world data, thus enhancing model robustness. 
Moreover, hybrid AI systems combining multimodal inputs, such as thermal imaging and video data, could provide richer contextual information to improve detection accuracy in complex settings.

%Seb: Voncarlos put a reference for the claim below
%Voncarlos:ok
While AI is well-adapted for monitoring animals' physiological indicators, it faces difficulties in interpreting their emotions and behaviors, an area beyond the scope of this study and intended for future exploration \cite{zhang2024advancements}.
Additionally, the successful deployment of AI technology necessitates continuous training and maintenance, requiring personnel with specialized expertise. 
To reduce dependence on specialized personnel, implementing automated retraining pipelines and explainable AI (XAI) methods could simplify model maintenance and interpretation, facilitating broader adoption even in facilities with limited technical expertise.

Furthermore, implementing AI systems, particularly in large facilities, often involves high initial costs, which may deter smaller organizations or individual proprietors \cite{Javaid2022}. 
Addressing this, future research could explore lightweight AI models optimized for edge computing to reduce hardware requirements and operational costs. 
These models, paired with decentralized processing architectures, can make AI systems more accessible to smaller farms.

Continuous AI surveillance of animals also raises ethical concerns regarding their privacy and autonomy \cite{Bossert2021}. 
Balancing the benefits of welfare monitoring with the need to respect animals' independence is essential. 
Therefore, collaborative research among technologists, veterinarians and animal behaviorists is crucial. 
This interdisciplinary approach ensures that AI technologies are designed and implemented in ways that truly benefit animals. 
By incorporating ethical AI frameworks, such as those ensuring transparency, accountability, and minimal invasiveness, researchers can build systems that prioritize animal well-being without compromising ethical standards.

In summary, the future of AI in animal welfare lies in developing multimodal systems, leveraging lightweight architectures, and addressing ethical challenges through interdisciplinary collaboration. These advancements will pave the way for scalable, cost-effective, and humane solutions that enhance both animal welfare and operational efficiency.

\section{Conclusion}
\label{sect_conc}
%Q.E.D.

This study offers a detailed evaluation of cow detection models using deep learning, with a focus on their performance in complex scenarios.
We compared YOLOv5, YOLOv8 and Mask R-CNN models across various conditions, employing Stochastic Gradient Descent (SGD) and Adam optimizers to improve training outcomes. 
The main challenges addressed include complex backgrounds, occlusions, variable lighting conditions and different camera perspectives, all of which complicate object detection.

Among the models tested, YOLOv8, particularly when enhanced with the CBAM module, demonstrated superior performance in handling diverse and challenging scenarios. 
YOLOv8's grid-based predictions and anchor boxes enabled it to manage varying cow sizes and complex scenes effectively. 
The YOLOv8-CBAM model achieved the highest precision and recall, showing remarkable robustness in detecting cows across both indoor and outdoor environments.

Integrating the CBAM attention mechanism into YOLOv8 addresses challenges such as overlapping cows, complex backgrounds and low-light conditions, significantly improving target saliency and reducing missed and false detections. 
The precision-recall (P-R) curves reveal that YOLOv8l-CBAM consistently outperforms YOLOv8l, with notable improvements in Average Precision (AP) across different camera types. 
While AP changes were minimal for IW, OC and OP cameras, indicating comparable performance between models, YOLOv8l-CBAM excelled with peak AP values of 99\% for OE and significant AP increases of 8.1\%, 2.8\%, and 1.7\% for respectively indoor cameras IC, IP and IW. 
These improvements are attributed to YOLOv8l-CBAM's optimized architecture, which effectively utilizes attention weights and regularization techniques to enhance performance even with limited data. 

%In summary, key contributions of this research include: (1) providing a detailed analysis of the limitations in cow detection in both challenging indoor and outdoor environments, (2) presenting a novel general model that effectively detects cows under complex real-world conditions, and (3) evaluating and benchmarking the latest detection algorithms to assess their performance.

This study underscores that while AI-based detection models can minimize the need for human intervention, they still face limitations. 
Indeed, solely relying on image data may overlook valuable insights from additional sources like GPS, RFID or thermal imaging. 
Future research should aim to integrate these data sources, develop hybrid models and explore ensemble techniques to further enhance detection performance and computational efficiency.

\bibliographystyle{cas-model2-names}

\bibliography{cas-refs}

\newpage

\end{document}